\documentclass[10pt,twocolumn]{article}

\usepackage[letterpaper,top=0.85in,bottom=1.05in,left=0.78in,right=0.78in,columnsep=0.25in]{geometry}
\usepackage{lmodern}
\usepackage[T1]{fontenc}
\usepackage{microtype}
\usepackage{graphicx}
\usepackage{amsmath}
\usepackage{amssymb}
\usepackage{booktabs}
\usepackage{multirow}
\usepackage{adjustbox}
\usepackage{caption}
\usepackage{subcaption}
\usepackage{enumitem}
\usepackage{algorithm}
\usepackage{algorithmic}
\usepackage{authblk}
\usepackage{xcolor}
\usepackage[colorlinks=true,linkcolor=blue!60!black,citecolor=blue!60!black,urlcolor=blue!60!black]{hyperref}

\newcommand{\name}{PhaseQuant}

\newcommand{\ie}{{\emph{i.e.,}\ }}

\title{\textbf{PhaseQuant: Mitigating Activation Quantization Collapse\\ via Residual-Stream Dynamics}}

\author[1]{Zimo Zhao\thanks{First author.}}
\author[1]{Maolin Wang}
\author[1]{Bowen Yu}
\author[1]{Bowen Liu}
\author[2]{Xiao Han\thanks{Corresponding author.}}
\author[1]{Xiangyu Zhao\thanks{Corresponding author.}}

\affil[1]{City University of Hong Kong, Hong Kong, China}
\affil[2]{Zhejiang University of Technology, Hangzhou, China}

\date{}

\begin{document}
\maketitle

\begin{abstract}
Post-training quantization (PTQ) is essential for efficient large language model (LLM) inference, but reliably quantizing activations remains challenging when weights, activations, and KV caches are all quantized to 4-bit precision. A key difficulty lies in massive activations, whose extreme values dominate the activation range and amplify quantization error. State-of-the-art methods mitigate massive activations through transformation-based smoothing, such as orthogonal rotations and affine scaling, but overlook the cross-layer dynamics of the residual stream. In this paper, we show that massive activations emerge and disappear in a phase-wise pattern across network depth, injecting abrupt updates into the residual stream. These updates cause the newly injected layer-wise signal to dominate the 4-bit quantization scale and weaken the accumulated residual information. To characterize this behavior, we introduce two metrics, the \emph{Update Dominance Ratio} (UDR) and the \emph{Historical Retention Score} (HRS), and show that static transformation-based smoothing cannot fully resolve the resulting dynamic quantization instability. Based on this analysis, we propose \textbf{\name}, a phase-aware mixed-precision post-training quantization policy for activation quantization. \name{} identifies quantization-sensitive layers from residual-stream dynamics and assigns 8-bit activation precision only to these layers, while keeping weights, KV caches, and all remaining activations at 4-bit precision. It can be directly integrated with strong PTQ baselines such as QuaRot, SpinQuant, and FlatQuant. Experiments on LLaMA-2 and LLaMA-3 show that \name{} consistently improves perplexity and zero-shot QA performance under W4A4KV4 quantization, while achieving $1.05\times$--$1.07\times$ throughput improvement with modest memory overhead. These results demonstrate a practical path toward robust low-bit LLM inference.
\end{abstract}

\noindent\textbf{Keywords:} Resource-Efficient AI; Large Language Models; Post-Training Quantization; Efficient LLM Inference; Model Compression

\section{Introduction}
\label{sec:introduction}

Large language models (LLMs) have demonstrated strong performance
across a broad range of tasks, including natural language understanding,
reasoning, and code generation, as their scale increases
rapidly~\cite{kaplan2020scaling,achiam2023gpt,grattafiori2024llama}.
However, larger models entail significantly higher memory and computational
demands, making efficient inference increasingly critical for practical
deployment. Meanwhile, achieving efficient inference further enables local LLM deployment, which is essential for satisfying strict data privacy requirements and
reducing latency in time-sensitive scenarios~\cite{zheng2025reviewedgelargelanguage}. To this end, post-training quantization (PTQ) has emerged as one of the most practical approaches, quantizing pretrained weights, activations, and KV caches into low-bit representations without retraining, thereby reducing memory footprint and enabling efficient low-precision inference~\cite{nagel2020up,xiao2023smoothquant,frantar2022gptq,sun2024flatquant}.

However, PTQ introduces \textbf{non-negligible quantization error} that degrades model performance, particularly at 4-bit precision. Empirical studies have revealed that weight distributions are relatively smooth and easy to quantize~\cite{xiao2023smoothquant}. In contrast, activations present a much greater challenge: within certain tokens, a few channels can exhibit activation values that are orders of magnitude larger than those of the remaining channels, a phenomenon known as \textbf{massive activations}~\cite{dettmers2022gpt3,queipo2025attention,sun2024massive}. These massive activations dominate the quantization range, squeezing the majority of normal values into very few representable levels and significantly increasing quantization error.

To mitigate this issue, early approaches like SmoothQuant~\cite{xiao2023smoothquant} and AWQ~\cite{lin2024awq} introduce per-channel scaling factors that shift quantization difficulty from activations to the more quantization-friendly weights. While effective at 8-bit precision, channel-wise scaling becomes insufficient at 4-bit precision. The sharply reduced representable levels fail to accommodate residual outliers, leading to severe quantization collapse. Recent methods move beyond local scaling by applying global transformations to the weight and activation distributions before quantization: QuaRot~\cite{ashkboos2024quarot} applies randomized Hadamard rotations, SpinQuant learns orthogonal rotations~\cite{liu2025spinquant}, and FlatQuant~\cite{sun2024flatquant} generalizes the transform to a learnable affine mapping for stronger distribution flattening. These transform-based PTQ methods currently achieve state-of-the-art performance under 4-bit precision settings, indicating that they can already provide accurate layer-wise activation quantization, where the updated activations often remain within the calibrated 4-bit quantization range.

Despite their differences, all of the above transforms share one structural assumption in PTQ activation quantization: \textbf{each Transformer layer is treated as a static and isolated unit}, with the transform calibrated once per layer on a small calibration set. Even under per-token activation quantization, the transform remains fixed at the layer level during inference. This static layer-wise view neglects how massive activations evolve across layers as the model processes inputs: certain tokens, such as the BOS token, produce massive activations, giving rise to two related phenomena, attention sinks and compression valleys~\cite{queipo2025attention}.

This mechanism partitions network depth into three phases. In \textbf{Phase 1}, early layers perform broad information mixing. In \textbf{Phase 2}, middle layers enter representational compression, where token representations become more compact and massive activations dominate the residual stream. In \textbf{Phase 3}, layers near the output perform selective refinement for the final prediction. The emergence or disappearance of massive activations marks the boundaries between these phases.

\begin{figure}[t]
    \centering
    \includegraphics[width=\linewidth]{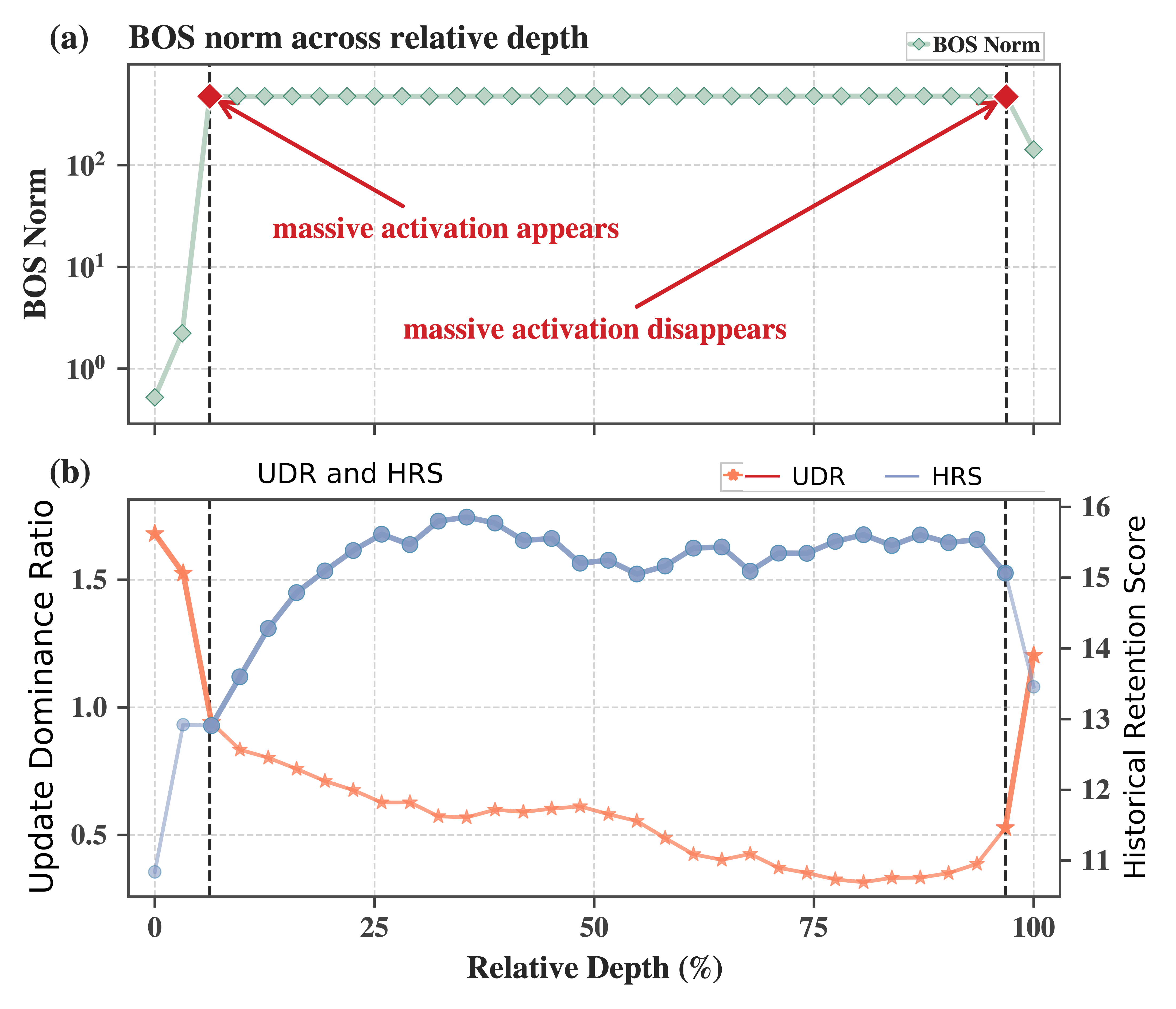}
    \caption{Massive activations and residual-stream dynamics under 4-bit activation quantization.
    (a) BOS token norm in Llama-3-8B, which identifies the emergence and disappearance of massive activations.
    (b) Update Dominance Ratio (UDR) and Historical Retention Score (HRS) under FlatQuant with 4-bit activation-only quantization.
    UDR quantifies the relative magnitude of the current residual update, while HRS measures the preservation of accumulated residual information.}
    \label{fig:layerwise_combined}
\end{figure}

Figure~\ref{fig:layerwise_combined} shows that the layer-wise evolution of massive activations is closely coupled with low-bit activation quantization errors.
The degradation is mainly concentrated in Phase 1 and Phase 3, where the emergence and disappearance of massive activations introduce large cross-layer norm shifts in the residual stream.
Under FlatQuant with 4-bit activation-only quantization, these two phases exhibit higher Update Dominance Ratios and lower Historical Retention Scores, suggesting that the dominant current update increases the per-token quantization step size and makes smaller historical information more vulnerable to rounding errors.

Motivated by this observation, we propose \textbf{\name}, a \textbf{phase-aware mixed-precision quantization strategy} that adapts activation precision to the residual-stream dynamics of each phase. Specifically, we assign 8-bit activation precision to Phase 1 and Phase 3, where larger residual updates require finer quantization to preserve historical information.
In contrast, Phase 2 has weaker and smoother residual updates, so 4-bit activation quantization remains sufficient while reducing inference cost. This phase-aware design, therefore, addresses the cross-layer heterogeneity that static PTQ methods usually overlook.
Our specific contributions are as follows:
\begin{itemize}[leftmargin=*]
\itemsep0em
\item We introduce two new metrics, the Update Dominance Ratio and the Historical Retention Score, to measure quantization loss in the residual stream, and theoretically show that quantization degradation at phase boundaries is unavoidable, even with orthogonal transforms.
\item Guided by this analysis, we propose a phase-aware mixed-precision activation policy that identifies sensitive layers from calibration-time residual-stream statistics and assigns 8-bit activations only to them, while keeping all other activations, weights, and KV caches in 4-bit precision.
\item Our strategy can be directly integrated with strong transform-based PTQ baselines (QuaRot, SpinQuant, and FlatQuant). Under 4-bit quantization on LLaMA-2 and LLaMA-3, it consistently improves these baselines and brings FlatQuant closer to the FP16 upper bound.
\end{itemize}

\section{Preliminaries}
\label{sec:preliminaries}

\paragraph{\textbf{Massive Activations.}}
Empirical studies have shown that Transformer residual streams often contain massive activations, frequently concentrated on special tokens such as the beginning-of-sequence (BOS) token~\cite{sun2024flatquant,queipo2025attention,sun2024massive}. From the perspective of residual flow, these activations appear as abrupt changes in norm: certain tokens acquire much larger residual norms than ordinary tokens and dominate the layer-wise state. This behavior follows a three-phase pattern across depth. In the early layers, massive activations emerge rapidly and introduce large residual updates, forming the first region of abrupt residual updates. In the middle compression-valley layers, the residual stream becomes more stable, and layer-wise updates remain small. Near the output layers, the residual norm rises again as massive activations are redistributed, producing another abrupt-update region. Thus, massive activations are not isolated layer-wise outliers, but structured cross-layer dynamics that are highly consistent across sequences and decoder-only model families.

\paragraph{\textbf{Low-Bit Quantization of LLMs.}}
PTQ is widely used to reduce the memory footprint and inference latency of large language models.
It replaces full-precision weights
$\boldsymbol{W}\in\mathbb{R}^{m\times n}$
and activations
$\boldsymbol{X}\in\mathbb{R}^{k\times n}$
in linear projections, \ie,
$\boldsymbol{Y}=\boldsymbol{X}\boldsymbol{W}^{\top}$,
with low-bit representations.
In autoregressive decoding, the key-value (KV) cache can also be quantized to reduce memory overhead further.

For a $b$-bit uniform quantizer, let
$\boldsymbol{x}\in\mathbb{R}^{d}$ denote the activation vector of one token.
The quantization step size is determined by its dynamic range:
\begin{equation}
s_b(\boldsymbol{x})
=
\frac{
\max(\boldsymbol{x})-\min(\boldsymbol{x})
}{
2^b-1
}.
\label{eq:quant_step}
\end{equation}
The quantized activation can be written as
\begin{equation}
Q_b(\boldsymbol{x})
=
s_b(\boldsymbol{x})
\left\lfloor
\frac{\boldsymbol{x}}{s_b(\boldsymbol{x})}
\right\rceil,
\label{eq:activation_quant}
\end{equation}
where $\lfloor\cdot\rceil$ denotes element-wise rounding.
The corresponding quantization error is
\begin{equation}
\boldsymbol{e}_b(\boldsymbol{x})
=
\boldsymbol{x}
-
Q_b(\boldsymbol{x}).
\label{eq:quant_error}
\end{equation}

This formulation highlights that the effective resolution of a $b$-bit quantizer is fundamentally determined by the step size $s_b$: a larger dynamic range inevitably yields a larger step size, leaving fewer distinguishable levels for small-magnitude features and thereby amplifying rounding error during quantization.

\paragraph{\textbf{Rotation-based Activation Transformation.}}
Recent PTQ methods commonly apply rotation-based transformations before low-bit activation quantization to mitigate the effect of massive activations. These transformations can be implemented by fixed orthogonal matrices, such as Hadamard rotations, or by learned orthogonal rotations. For a unified formulation, we denote the rotation matrix at layer $\ell$ as $\boldsymbol{R}^{(\ell)}\in\mathbb{R}^{d\times d}$, with $\boldsymbol{R}^{(\ell)\top}\boldsymbol{R}^{(\ell)}=\boldsymbol{I}$. Given the activation $\boldsymbol{X}^{(\ell)}$ at layer $\ell$, quantization is performed in the rotated space:
\begin{equation}
\widehat{\boldsymbol{X}}^{(\ell)} = Q_b\!\left(\boldsymbol{X}^{(\ell)}\boldsymbol{R}^{(\ell)}\right) \boldsymbol{R}^{(\ell)\top}.
\label{eq:rotation_activation}
\end{equation}
The rotation redistributes the magnitude of massive activations across hidden channels. Since per-token activation quantization determines the quantization step size based on the token-wise dynamic range, a few highly concentrated, massive activations can enlarge the range and make the low-bit grid coarse across most feature dimensions. By spreading large activation magnitudes more evenly across channels, rotation-based transformations produce a smoother activation range and reduce rounding error under low-bit quantization. For a linear projection, the rotation can be absorbed into the weight:
\begin{equation}
\boldsymbol{X}^{(\ell)}\boldsymbol{W}^{(\ell)\top} = \left(\boldsymbol{X}^{(\ell)}\boldsymbol{R}^{(\ell)}\right) \left(\boldsymbol{W}^{(\ell)}\boldsymbol{R}^{(\ell)}\right)^{\top}.
\label{eq:rotation_equiv}
\end{equation}
Thus, for each layer, $\boldsymbol{W}^{(\ell)}\boldsymbol{R}^{(\ell)}$ can be precomputed offline, while the rotation remains fixed during inference. Activations are therefore quantized in a smoother but static layer-wise rotated space. This preserves the original linear mapping and reduces quantization error caused by channel-wise massive activations.

\paragraph{\textbf{Problem Formulation.}}
We study activation quantization in decoder-only Transformers from the perspective of residual-stream dynamics.
Let $\boldsymbol{x}^{(\ell)}_t \in \mathbb{R}^{d}$ denote the residual state of token $t$ entering layer $\ell$.
Each layer applies an attention-MLP transformation $\mathcal{F}^{(\ell)}(\cdot)$ and updates the residual stream as
\begin{equation}
\boldsymbol{x}^{(\ell+1)}_t
=
\boldsymbol{x}^{(\ell)}_t
+
\Delta \boldsymbol{x}^{(\ell)}_t,
\qquad
\Delta \boldsymbol{x}^{(\ell)}_t
=
\mathcal{F}^{(\ell)}
\left(
\boldsymbol{x}^{(\ell)}
\right)_t .
\label{eq:residual_transition}
\end{equation}
Here, $\boldsymbol{x}^{(\ell)}_t$ represents the historical residual information accumulated from previous layers, while $\Delta \boldsymbol{x}^{(\ell)}_t$ denotes the new layer-wise update.

With $b$-bit activation quantization, the residual update becomes
\begin{equation}
\widehat{\boldsymbol{x}}^{(\ell+1)}_t
=
Q_b
\left(
\widehat{\boldsymbol{x}}^{(\ell)}_t
+
\Delta \boldsymbol{x}^{(\ell)}_t
\right),
\qquad
\widehat{\boldsymbol{x}}^{(0)}_t
=
\boldsymbol{x}^{(0)}_t .
\label{eq:quantized_residual_transition}
\end{equation}
This shows that the same low-bit quantization grid must represent both the historical residual information and the current layer update.
When massive activations introduce abrupt residual updates, the current update can dominate the quantization scale, making the smaller historical residual information more vulnerable to rounding error.

\section{Method}
\label{sec:method}

\begin{figure}[t]
    \centering
    \includegraphics[width=\columnwidth]{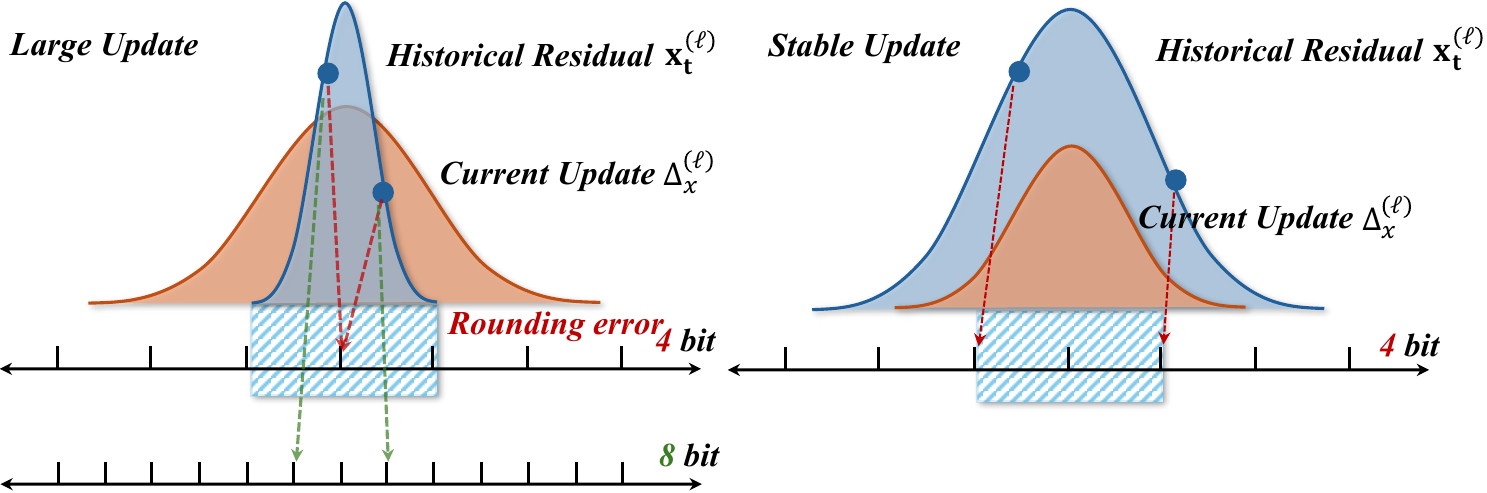}
    \vspace{-0.6em}
    \caption{Illustration of residual-stream dynamics under low-bit activation quantization.}
    \label{fig:residual_stream_dynamics}
    \vspace{-0.8em}
\end{figure}

Based on the residual-stream dynamics described in
Eq.~\eqref{eq:residual_transition}, we study activation quantization through the
interaction between the historical residual state
$\boldsymbol{x}^{(\ell)}_t$ and the current layer update
$\Delta \boldsymbol{x}^{(\ell)}_t$ across Transformer depth.
Existing PTQ methods mainly treat each layer as an independent quantization unit.
They reduce activation error by rescaling or rotating activations to suppress
channel-wise outliers. However, such a static view does not model the cross-layer
change in the relative scale between historical residual information and newly
injected updates.

Our key observation is that 4-bit activation degradation is not uniformly
distributed across layers, but is closely coupled with residual-stream dynamics.
Following the phase behavior discussed in the preliminaries, 4-bit activation
degradation mainly appears in Phase~1 and Phase~3, where the emergence and disappearance of massive activations introduce large residual updates.
Figure~\ref{fig:residual_stream_dynamics} illustrates why these large-update layers are more sensitive to 4-bit quantization. At these layers, the current
update $\Delta \boldsymbol{x}^{(\ell)}_t$ can dominate the historical residual
state $\boldsymbol{x}^{(\ell)}_t$. The 4-bit quantizer then tends to fit this
dominant update by expanding its effective range, which coarsens the grid for
smaller historical residual information. Consequently, historical semantic
features occupy fewer quantization intervals and become more vulnerable to
rounding error.

In contrast, Phase~2 corresponds to a compression valley where residual updates
remain stable and moderate. The current update does not dominate the historical
residual state. Thus, 4-bit activations still provide sufficient resolution to
preserve the main semantic variations, while avoiding the additional memory and
inference cost of applying 8-bit activations to all layers.

Based on this observation, we use two residual-dynamics metrics to identify
precision-sensitive layers: the \emph{Update Dominance Ratio} (UDR), which measures the relative
strength of the current update, and the \emph{Historical Retention Score} (HRS), which measures the
preservation of historical residual information after quantization. Layers
with dominant updates and degraded historical preservation are assigned
8-bit activation precision, while the remaining layers stay in 4-bit precision
for efficient inference.

\subsection{Update Dominance Ratio and Historical Retention Score}
\label{subsec:udr_hrs}

Eq.~\eqref{eq:quantized_residual_transition} shows that activation quantization
is applied to the updated residual state
$\widehat{\boldsymbol{x}}^{(\ell)}_t+\Delta\boldsymbol{x}^{(\ell)}_t$.
Thus, the same low-bit grid must represent both the historical residual state
$\widehat{\boldsymbol{x}}^{(\ell)}_t$ and the newly injected update
$\Delta\boldsymbol{x}^{(\ell)}_t$.

For per-token activation quantization, the step size is determined by the
token-wise dynamic range, as defined in Eq.~\eqref{eq:quant_step}. When the
current update is large, it can dominate the dynamic range of the combined state
and enlarge the 4-bit step size. Even if the updated activation does not exceed
the valid quantization range, the enlarged step size reduces the effective
resolution for smaller historical residual features. This makes them more
vulnerable to rounding error and explains why activation degradation tends to
concentrate in large-update depth regions.

We measure the strength of this residual update using the Update Dominance Ratio:
\begin{equation}
\mathrm{UDR}_{\ell,t}
=
\frac{
\left\|
\Delta \boldsymbol{x}^{(\ell)}_t
\right\|_2
}{
\left\|
\boldsymbol{x}^{(\ell)}_t
\right\|_2
+
\varepsilon
},
\label{eq:udr_token}
\end{equation}
where $\varepsilon$ is a small constant for numerical stability. The layer-level
UDR is
\begin{equation}
\mathrm{UDR}_{\ell}
=
\frac{1}{T}
\sum_{t=1}^{T}
\mathrm{UDR}_{\ell,t}.
\label{eq:udr_layer}
\end{equation}
A larger $\mathrm{UDR}_{\ell}$ indicates that the current layer introduces a
stronger update relative to the historical residual stream.

However, a large UDR alone does not fully determine whether 4-bit
activation quantization is harmful. It measures the relative scale of the
current update, but does not directly show whether historical residual
information is distorted after quantization. We therefore measure how much
historical residual information is preserved after quantization. Given the
quantized updated residual state
$\widehat{\boldsymbol{x}}^{(\ell+1)}_t =
Q_b(\boldsymbol{x}^{(\ell)}_t+\Delta \boldsymbol{x}^{(\ell)}_t)$, and since
$\Delta\boldsymbol{x}^{(\ell)}_t$ is the current layer update, the historical
residual reconstructed from the quantized activation is
\begin{equation}
\widehat{\boldsymbol{x}}^{(\ell)}_t
=
\widehat{\boldsymbol{x}}^{(\ell+1)}_t
-
\Delta \boldsymbol{x}^{(\ell)}_t .
\label{eq:reconstructed_history}
\end{equation}
We then define the Historical Retention Score as
\begin{equation}
\mathrm{HRS}_{\ell}
=
\frac{1}{T}
\sum_{t=1}^{T}
10\log_{10}
\frac{
\left\|
\boldsymbol{x}^{(\ell)}_t
\right\|_2^2
}{
\left\|
\widehat{\boldsymbol{x}}^{(\ell)}_t
-
\boldsymbol{x}^{(\ell)}_t
\right\|_2^2
+
\varepsilon
}.
\label{eq:hrs_def}
\end{equation}
A higher $\mathrm{HRS}_{\ell}$ indicates better preservation of
historical residual information, while a lower value indicates stronger
quantization-induced distortion.

\subsection{Limits of Static Rotation Smoothing}
\label{subsec:rotation_limitation}

Rotation-based transformations are effective for smoothing massive activations
across hidden channels. However, they are static layer-wise transformations and
do not explicitly change cross-layer residual dynamics. Let
$\boldsymbol{R}^{(\ell)}$ be the orthogonal rotation matrix used at layer
$\ell$, satisfying
$\boldsymbol{R}^{(\ell)\top}\boldsymbol{R}^{(\ell)}=\boldsymbol{I}$.
Applying the same rotation to the historical residual and the current update
preserves their Euclidean norms:
\begin{equation}
\left\|
\boldsymbol{x}^{(\ell)}_t\boldsymbol{R}^{(\ell)}
\right\|_2
=
\left\|
\boldsymbol{x}^{(\ell)}_t
\right\|_2,
\qquad
\left\|
\Delta\boldsymbol{x}^{(\ell)}_t\boldsymbol{R}^{(\ell)}
\right\|_2
=
\left\|
\Delta\boldsymbol{x}^{(\ell)}_t
\right\|_2 .
\label{eq:rotation_norm}
\end{equation}
Therefore, the relative strength of the current update remains invariant under
the rotation:
\begin{equation}
\frac{
\left\|
\Delta\boldsymbol{x}^{(\ell)}_t\boldsymbol{R}^{(\ell)}
\right\|_2
}{
\left\|
\boldsymbol{x}^{(\ell)}_t\boldsymbol{R}^{(\ell)}
\right\|_2+\varepsilon
}
=
\frac{
\left\|
\Delta\boldsymbol{x}^{(\ell)}_t
\right\|_2
}{
\left\|
\boldsymbol{x}^{(\ell)}_t
\right\|_2+\varepsilon
}
=
\mathrm{UDR}_{\ell,t}.
\label{eq:udr_rotation_invariant}
\end{equation}

This invariance shows that rotation-based smoothing can reduce channel-wise
massive activation effects, but it does not remove the abrupt residual update
itself.
In other words, the rotation smooths the activation distribution within each
token, but it cannot reduce the relative scale of the newly injected update with
respect to the historical residual state. As a result, even after static
rotation-based smoothing, phase-boundary layers may remain sensitive to 4-bit
activation quantization.

We further analyze how the remaining abrupt residual update affects historical
information preservation. Consider activation quantization in the rotated space:
\begin{equation}
\begin{aligned}
\widehat{\boldsymbol{x}}^{(\ell+1)}_{R,t}
&=
Q_b
\left(
\left(
\boldsymbol{x}^{(\ell)}_t
+
\Delta \boldsymbol{x}^{(\ell)}_t
\right)
\boldsymbol{R}^{(\ell)}
\right)
\\
&=
\left(
\boldsymbol{x}^{(\ell)}_t
+
\Delta \boldsymbol{x}^{(\ell)}_t
\right)
\boldsymbol{R}^{(\ell)}
+
\boldsymbol{\epsilon}^{(\ell)}_t ,
\end{aligned}
\label{eq:rotation_quant_noise}
\end{equation}
where $\boldsymbol{\epsilon}^{(\ell)}_t$ denotes the quantization noise in the
rotated space. Mapping the quantized state back to the original residual space
gives
\begin{equation}
\begin{aligned}
\widehat{\boldsymbol{x}}^{(\ell+1)}_t
&=
\widehat{\boldsymbol{x}}^{(\ell+1)}_{R,t}
\boldsymbol{R}^{(\ell)\top}
=
\boldsymbol{x}^{(\ell)}_t
+
\Delta \boldsymbol{x}^{(\ell)}_t
+
\boldsymbol{\delta}^{(\ell)}_t,
\\
\boldsymbol{\delta}^{(\ell)}_t
&=
\boldsymbol{\epsilon}^{(\ell)}_t
\boldsymbol{R}^{(\ell)\top}.
\end{aligned}
\label{eq:inverse_rotation_error}
\end{equation}
By orthogonality, the inverse transformation preserves the error norm:
\begin{equation}
\left\|
\boldsymbol{\delta}^{(\ell)}_t
\right\|_2
=
\left\|
\boldsymbol{\epsilon}^{(\ell)}_t
\right\|_2 .
\label{eq:error_norm_preserve}
\end{equation}
Therefore, the inverse rotation does not remove quantization noise; it only maps
the noise back to the original residual space.

Under the standard uniform quantization noise model, the expected noise energy
scales with the square of the quantization step size:
\begin{equation}
\mathbb{E}
\left[
\left\|
\boldsymbol{\epsilon}^{(\ell)}_t
\right\|_2^2
\right]
=
\Theta(d\,s_b^2),
\label{eq:noise_energy_step}
\end{equation}
where $s_b$ denotes the token-wise quantization step size. When the current
update dominates the updated residual state, the token-wise dynamic range, and
therefore $s_b$, is largely determined by
$\Delta \boldsymbol{x}^{(\ell)}_t$. This gives the following scaling trend:
\begin{equation}
\mathbb{E}
\left[
\left\|
\boldsymbol{\delta}^{(\ell)}_t
\right\|_2^2
\right]
=
\mathbb{E}
\left[
\left\|
\boldsymbol{\epsilon}^{(\ell)}_t
\right\|_2^2
\right]
\propto
\frac{
\left\|
\Delta \boldsymbol{x}^{(\ell)}_t
\right\|_2^2
}{
2^{2b}
}.
\label{eq:error_update_scale}
\end{equation}

Since the historical signal to be preserved is
$\boldsymbol{x}^{(\ell)}_t$, the Historical Retention Score follows
\begin{equation}
\mathrm{HRS}_{\ell,t}
\propto
\frac{
\left\|
\boldsymbol{x}^{(\ell)}_t
\right\|_2^2
}{
\mathbb{E}
\left[
\left\|
\boldsymbol{\delta}^{(\ell)}_t
\right\|_2^2
\right]
}
\approx
\mathcal{O}
\left(
\frac{
2^{2b}
}{
\mathrm{UDR}_{\ell,t}^{2}
}
\right).
\label{eq:hrs_udr_inverse}
\end{equation}

Equation~\eqref{eq:hrs_udr_inverse} suggests that a larger update dominance
ratio reduces the effective preservation of historical residual information under
low-bit activation quantization. We emphasize that this relation is a scaling
trend rather than an exact equality, since the realized quantization error also
depends on the activation distribution and quantizer implementation. Therefore, the UDR alone is not sufficient for layer selection. We instead combine the UDR with the HRS, where the UDR identifies layers with strong residual updates, and the HRS measures the resulting degradation of historical residual information after 4-bit quantization.

\subsection{Phase-Aware Mixed-Precision Activation Policy}
\label{subsec:layer_selection}

The above analysis suggests that higher activation precision should be assigned
to layers where 4-bit activation quantization fails to preserve historical
residual information. We refer to these layers as low-bit
quantization-sensitive layers. To characterize such layers, we use two
residual-dynamics signals: the UDR and the HRS. The sensitive
layer set can be written as
\begin{equation}
\mathcal{S}
=
\left\{
\ell:
\mathrm{UDR}_{\ell}>\tau_{\mathrm{UDR}}
\;\wedge\;
\mathrm{HRS}_{\ell}<\tau_{\mathrm{HRS}}
\right\}.
\label{eq:sensitive_layers}
\end{equation}
Here, $\ell$ denotes the layer index, $\mathrm{UDR}_{\ell}$ measures the relative
strength of the residual update at layer $\ell$, and
$\mathrm{HRS}_{\ell}$ measures how well historical residual
features are preserved after 4-bit activation quantization.

Our calibration analysis shows a consistent depth-wise pattern in dense
decoder-only LLMs: the first few layers and the final layer are more vulnerable
to historical residual information loss under 4-bit activation quantization.
Accordingly, for an $L$-layer dense decoder-only model, we instantiate the
low-bit quantization-sensitive layer set as
\begin{equation}
\mathcal{S}
=
\{1,2,3,L\},
\label{eq:sensitive_set}
\end{equation}
where layers are indexed from $1$ to $L$. Given $\mathcal{S}$, \name{} applies
phase-aware precision allocation by assigning higher activation precision only to
these sensitive layers, while keeping weights, non-sensitive activations, and KV
caches in 4-bit precision. Algorithm~\ref{alg:phasequant} summarizes the overall
workflow.

\begin{algorithm}[t]
\caption{\name}
\label{alg:phasequant}
\small
\begin{algorithmic}[1]
\REQUIRE Pretrained decoder-only model $\mathcal{M}$ with $L$ Transformer layers;
PTQ backbone $\mathcal{B}$; calibration data $\mathcal{D}_{\mathrm{cal}}$;
sensitive layer set $\mathcal{S}$
\ENSURE Quantized model $\widehat{\mathcal{M}}$

\STATE Set the weight and KV-cache precision to 4-bit:
\[
b_w=b_k=b_v=4 .
\]

\STATE Set the activation precision according to the phase-aware policy:
\[
b_x^{(\ell)}=
\begin{cases}
8, & \ell\in\mathcal{S},\\
4, & \ell\notin\mathcal{S},
\end{cases}
\quad \ell=1,\ldots,L .
\]

\STATE Apply the PTQ backbone $\mathcal{B}$ to transform $\mathcal{M}$ into a
quantization-friendly form:
\[
\mathcal{M}^{\prime}\leftarrow \Phi_{\mathcal{B}}(\mathcal{M}) .
\]

\STATE Quantize weights, activations, and KV caches using the assigned bit-widths:
\[
\mathcal{M}^{\prime}
\leftarrow
Q_{\mathcal{B}}
\left(
\mathcal{M}^{\prime};
b_w,\{b_x^{(\ell)}\}_{\ell=1}^{L},b_k,b_v
\right).
\]

\STATE Estimate the quantization parameters on calibration data:
\[
\Theta_Q
\leftarrow
\operatorname{Calibrate}_{\mathcal{B}}
\left(
\mathcal{M}^{\prime},
\mathcal{D}_{\mathrm{cal}};
b_w,\{b_x^{(\ell)}\}_{\ell=1}^{L},b_k,b_v
\right).
\]

\STATE Build the final quantized model using the calibrated parameters:
\[
\widehat{\mathcal{M}}
\leftarrow
Q_{\mathcal{B}}
\left(
\mathcal{M}^{\prime};
\Theta_Q
\right).
\]

\STATE \textbf{return} $\widehat{\mathcal{M}}$
\end{algorithmic}
\end{algorithm}

In Algorithm~\ref{alg:phasequant}, $\Theta_Q$ denotes the calibrated
quantization parameters, including step sizes, zero-points, clipping ranges, and
backbone-specific transformation parameters when applicable.

\paragraph{\textbf{Effect of phase-aware precision allocation.}}
The benefit of assigning higher precision to sensitive layers can be understood
from the quantization step size. For a fixed token-wise dynamic range, increasing
the activation bit-width from $b_{\mathrm{low}}$ to $b_{\mathrm{high}}$ reduces
the step size approximately as
\begin{equation}
\frac{s_{b_{\mathrm{high}}}}{s_{b_{\mathrm{low}}}}
\approx
2^{-(b_{\mathrm{high}}-b_{\mathrm{low}})} .
\label{eq:precision_step_ratio}
\end{equation}
Since the variance of uniform quantization error scales with the squared step
size, moving from 4-bit to 8-bit activation quantization reduces the error
variance by roughly $2^{-8}$ under the same dynamic range. Therefore, using
higher activation precision at sensitive layers provides a finer quantization
grid for preserving historical residual features. Meanwhile, the efficiency
benefit of low-bit inference is largely retained, because only a small subset of
activations is assigned 8-bit precision, while the remaining activations, all
weights, and all KV caches stay in 4-bit precision.

\section{Experiments}
\label{sec:experiments}

\subsection{Evaluation and Baselines}
Our main experiments cover dense decoder-only LLMs from the LLaMA-2~\cite{touvron2023llama} and LLaMA-3~\cite{grattafiori2024llama} series. We also analyze dense and mixture-of-experts decoder-only model architectures from the perspective of dynamic residual-stream behavior, with details provided in Section~\ref{sec:moe_residual_dynamics}.

For language modeling evaluation, we report perplexity (PPL) on WikiText-2~\cite{merity2016pointer} and C4~\cite{raffel2020exploring}.
For commonsense reasoning evaluation, we use six zero-shot tasks: ARC-Challenge, ARC-Easy~\cite{clark2018think}, HellaSwag~\cite{zellers2019hellaswag}, LAMBADA~\cite{paperno2016lambada}, PIQA~\cite{bisk2020piqa}, and WinoGrande~\cite{sakaguchi2021winogrande}.

We apply the \textbf{\name} policy to strong W4A4KV4 rotation-based activation transformation PTQ baselines, including QuaRot~\cite{ashkboos2024quarot}, SpinQuant~\cite{liu2025spinquant}, and FlatQuant~\cite{sun2024flatquant}. This setting quantizes weights, activations, and KV cache to 4-bit precision. Among these baselines, FlatQuant serves as the strongest baseline under W4A4KV4 quantization.

\subsection{Implementation and Quantization Details}
\label{subsec:implementation_details}
All experiments use the same calibration split across methods to ensure a controlled comparison. The calibration set is sampled from WikiText-2~\cite{merity2016pointer}, with each sequence truncated or padded to 2048 tokens. The evaluation pipeline is kept fixed for all baselines, varying only the quantization method under test.

We evaluate all methods under the W4A4KV4 setting, where weights, activations, and KV cache are quantized to 4-bit precision by default. Weights are quantized using per-channel symmetric quantization, whereas activations employ per-token symmetric quantization. KV-cache quantization follows group-wise quantization with a group size of 128, consistent with common low-bit LLM inference practices~\cite{liu2024kivi,hooper2024kvquant,sun2024flatquant}. For weight quantization, we report results with both RTN and GPTQ~\cite{frantar2022gptq}. RTN rounds full-precision weights directly to the low-bit grid, while GPTQ uses calibration data to compensate for quantization errors via closed-form updates.

Under this unified setup, we evaluate QuaRot, SpinQuant, and FlatQuant as PTQ baselines.
QuaRot applies randomized Hadamard rotations, SpinQuant uses learned orthogonal rotations, and FlatQuant learns affine transformations.
QuaRot requires no additional training for its transformation parameters, whereas SpinQuant is evaluated with the released optimized rotation checkpoints without further training.
For FlatQuant, we train the affine transformations for 15 epochs on 128 WikiText-2 sequences of length 2048, using a batch size of 4.
The learning rates are set to $5\times10^{-3}$ for transformation parameters and $5\times10^{-2}$ for learnable clipping parameters.
Additional implementation details are provided in our released code.

\textbf{\name} is applied to these PTQ baselines without modifying their weight or KV-cache quantization. Instead of using uniform 4-bit activations across all layers, \textbf{\name} adopts a phase-aware mixed-precision activation policy, assigning 8-bit precision only to quantization-sensitive layers while keeping all remaining layers at 4-bit. Specifically, layers $\{1,2,3,32\}$ are assigned 8-bit activations for LLaMA-2-7B and LLaMA-3-8B, layers $\{1,2,3,40\}$ for LLaMA-2-13B, and layers $\{1,2,3,80\}$ for LLaMA-2-70B and LLaMA-3-70B. This design isolates the contribution of the dynamic activation precision strategy: all methods share the same model, calibration data, weight quantizer, and KV-cache quantization protocol, differing only in the assignment of activation precision across layers.

\begin{table*}[t]
\centering
\caption{WikiText-2 and C4 perplexity of LLaMA models under 4-bit weight, activation, and KV-cache quantization.
For each PTQ baseline, \textbf{\name} keeps weights and KV caches at 4-bit, while using 8-bit activations only at phase-boundary layers. \textbf{Bold} indicates the best result for each model and dataset.}
\label{tab:llama4bit_phasequant}
\scriptsize
\renewcommand{\arraystretch}{1.08}
\setlength{\tabcolsep}{3pt}

\resizebox{\textwidth}{!}{%
\begin{tabular}{l c c c c c c c c c c c}
\toprule
\textbf{Method}
& \textbf{W Quant.}
& \multicolumn{5}{c}{\textbf{WikiText-2}}
& \multicolumn{5}{c}{\textbf{C4}} \\
\cmidrule(lr){3-7} \cmidrule(lr){8-12}
&
& \textbf{2-7B}
& \textbf{2-13B}
& \textbf{2-70B}
& \textbf{3-8B}
& \textbf{3-70B}
& \textbf{2-7B}
& \textbf{2-13B}
& \textbf{2-70B}
& \textbf{3-8B}
& \textbf{3-70B} \\
\midrule

FP16
& -
& 5.47 & 4.88 & 3.32 & 6.14 & 2.86
& 7.26 & 6.73 & 5.71 & 9.45 & 7.17 \\

\midrule

QuaRot
& RTN
& 8.56 & 6.10 & 4.14 & 10.60 & NA
& 11.86 & 8.67 & 6.42 & 17.19 & NA \\

QuaRot + \textbf{\name}
& RTN
& 7.97
& 6.06
& 4.11
& 9.56
& NA
& 11.10
& 8.60
& 6.40
& 15.28
& NA \\

SpinQuant
& RTN
& 6.14 & 5.44 & 3.82 & 7.96 & 7.58
& 9.19 & 8.11 & 6.26 & 13.45 & 15.39 \\

SpinQuant + \textbf{\name}
& RTN
& 6.06
& 5.41
& 3.81
& 7.70
& 7.52
& 9.05
& 8.03
& 6.23
& 12.92
& 15.36 \\

FlatQuant
& RTN
& 5.79 & 5.12 & 3.55 & 6.98 & 3.78
& 7.79 & 7.09 & 5.91 & 11.13 & 7.86 \\

FlatQuant + \textbf{\name}
& RTN
& \textbf{5.63}$^{*}$
& \textbf{5.00}$^{*}$
& \textbf{3.45}$^{*}$
& \textbf{6.54}$^{*}$
& \textbf{3.37}$^{*}$
& \textbf{7.55}$^{*}$
& \textbf{6.93}$^{*}$
& \textbf{5.82}$^{*}$
& 10.33
& \textbf{7.58}$^{*}$ \\

\midrule

QuaRot
& GPTQ
& 6.10 & 5.40 & 3.79 & 8.16 & 6.60
& 8.32 & 7.54 & 6.12 & 13.38 & 12.87 \\

QuaRot + \textbf{\name}
& GPTQ
& 5.97
& 5.34
& 3.77
& 7.68
& 6.03
& 8.18
& 7.46
& 6.10
& 12.54
& 11.58 \\

SpinQuant
& GPTQ
& 5.96 & 5.24 & 3.70 & 7.39 & 6.21
& 8.28 & 7.48 & 6.07 & 12.19 & 12.82 \\

SpinQuant + \textbf{\name}
& GPTQ
& 5.88
& 5.22
& 3.69
& 7.19
& 6.08
& 8.12
& 7.41
& 6.05
& 11.81
& 12.78 \\

FlatQuant
& GPTQ
& 5.78 & 5.11 & 3.54 & 6.90 & 3.77
& 7.86 & 7.11 & 5.92 & 11.21 & 7.93 \\

FlatQuant + \textbf{\name}
& GPTQ
& 5.64
& 5.01
& 3.54
& 6.55
& 3.75
& \textbf{7.55}$^{*}$
& \textbf{6.93}$^{*}$
& 5.92
& \textbf{10.32}$^{*}$
& 7.89 \\

\bottomrule
\end{tabular}%
}

\vspace{0.3em}
\small\textit{Note.} Lower perplexity is better. ``*'' indicates statistically significant improvements (\ie, two-sided $t$-test with $p<0.05$) over the best baseline.
\end{table*}

\subsection{Results on Language Modeling Perplexity}

Table~\ref{tab:llama4bit_phasequant} reports the WikiText-2 and C4 perplexities of LLaMA models under W4A4KV4 quantization. Overall, \textbf{\name} consistently reduces perplexity across model families, datasets, PTQ baselines, and weight quantizers, indicating that the proposed phase-aware activation precision policy is complementary to existing rotation-based activation transformation methods. The improvements are most pronounced when the original quantized model is less stable. For example, under RTN weight quantization, QuaRot on LLaMA-3-8B obtains perplexities of $10.60$ on WikiText-2 and $17.19$ on C4, while applying \textbf{\name} reduces them to $9.56$ and $15.28$, respectively. The gains are smaller but still consistent for stronger baselines such as FlatQuant, where LLaMA-3-8B improves from $6.98$ to $6.54$ on WikiText-2 and from $11.13$ to $10.33$ on C4. This trend suggests that \textbf{\name} brings larger benefits when the activation quantization error is more severe, while still providing additional gains on already strong PTQ baselines.

The results also show that the magnitude of improvement depends on both model scale and dataset difficulty. In general, smaller or more fragile quantized models benefit more from \textbf{\name}, whereas larger models with stronger baseline perplexity tend to show smaller absolute gains. For instance, on LLaMA-3-8B with QuaRot-RTN, \textbf{\name} reduces C4 perplexity by $1.91$, while on larger and stronger settings the gains become more moderate. A similar pattern appears across datasets: improvements are often larger on C4 than on WikiText-2, suggesting that more diverse open-domain text is more sensitive to activation-side quantization errors. These observations support the purpose of our design: \textbf{\name} is not intended to replace existing PTQ transformations, but to selectively correct the layers where uniform 4-bit activation quantization becomes the main bottleneck.

An interesting observation is that after applying \textbf{\name}, RTN can match or even outperform GPTQ in several strong-baseline settings. This trend is most visible for FlatQuant. For example, on LLaMA-3-70B, FlatQuant with RTN and \textbf{\name} achieves $3.37$ on WikiText-2 and $7.58$ on C4, outperforming its GPTQ counterpart with $3.75$ and $7.89$. A similar case appears on LLaMA-2-70B C4, where FlatQuant-RTN with \textbf{\name} obtains $5.82$, compared with $5.92$ for FlatQuant-GPTQ with \textbf{\name}. This does not indicate that GPTQ is generally inferior to RTN. Instead, it suggests that once the activation-side bottleneck is alleviated, the advantage of calibration-aware weight reconstruction may become less decisive. Under W4A4KV4 quantization, especially when combined with strong activation transformation baselines, carefully allocating activation precision can be as important as improving weight-side reconstruction.

\begin{table*}[t]
\centering
\caption{Zero-shot QA task results of LLaMA models using FlatQuant as the PTQ baseline under 4-bit weight, activation, and KV-cache quantization.
\name{} keeps weights and KV caches at 4-bit, while using 8-bit activations only at phase-boundary layers. \textbf{Bold} indicates the best average result for each model.}
\label{tab:qa_4bit_phasequant_flatquant}
\scriptsize
\renewcommand{\arraystretch}{1.05}
\setlength{\tabcolsep}{3.2pt}
\resizebox{\textwidth}{!}{%
\begin{tabular}{l l c c c c c c c c}
\toprule
\textbf{Model}
& \textbf{Method}
& \textbf{W Quant.}
& \textbf{ARC-C}
& \textbf{ARC-E}
& \textbf{HellaSwag}
& \textbf{LAMBADA}
& \textbf{PIQA}
& \textbf{WinoGrande}
& \textbf{Avg.} \\
\midrule

\multirow{5}{*}{2-7B}
& FP16 & -- & 46.16 & 74.54 & 75.98 & 73.92 & 79.05 & 69.06 & 69.79 \\
\cmidrule(lr){2-10}
& FlatQuant & RTN & 43.26 & 72.05 & 73.64 & 72.04 & 77.26 & 69.53 & 67.96 \\
& FlatQuant + \name & RTN
& 43.94 & 73.23 & 74.58 & 73.34 & 77.80 & 67.88 & \textbf{68.46}$^{*}$ \\
& FlatQuant & GPTQ & 43.00 & 71.21 & 73.31 & 72.06 & 77.53 & 67.72 & 67.47 \\
& FlatQuant + \name & GPTQ
& 43.86 & 73.36 & 74.49 & 73.16 & 78.29 & 67.56 & 68.45 \\

\midrule

\multirow{5}{*}{2-13B}
& FP16 & -- & 49.15 & 77.44 & 79.39 & 76.73 & 80.47 & 72.14 & 72.55 \\
\cmidrule(lr){2-10}
& FlatQuant & RTN & 48.04 & 76.64 & 77.59 & 76.60 & 79.38 & 70.24 & 71.42 \\
& FlatQuant + \name & RTN
& 48.04 & 76.94 & 78.56 & 76.87 & 79.38 & 71.67 & \textbf{71.91}$^{*}$ \\
& FlatQuant & GPTQ & 48.38 & 76.94 & 77.88 & 76.40 & 79.65 & 70.56 & 71.64 \\
& FlatQuant + \name & GPTQ
& 48.21 & 77.57 & 78.37 & 76.79 & 79.71 & 70.72 & 71.89 \\

\midrule

\multirow{5}{*}{2-70B}
& FP16 & -- & 57.17 & 81.02 & 83.81 & 79.60 & 82.70 & 77.98 & 77.05 \\
\cmidrule(lr){2-10}
& FlatQuant & RTN & 56.14 & 80.30 & 83.01 & 79.60 & 82.75 & 77.90 & 76.62 \\
& FlatQuant + \name & RTN
& 57.00 & 80.68 & 83.38 & 80.07 & 81.83 & 77.98 & \textbf{76.82}$^{*}$ \\
& FlatQuant & GPTQ & 56.40 & 80.09 & 82.91 & 80.01 & 82.92 & 76.87 & 76.53 \\
& FlatQuant + \name & GPTQ
& 56.51 & 80.16 & 83.04 & 80.08 & 83.11 & 76.89 & 76.62 \\

\midrule

\multirow{5}{*}{3-8B}
& FP16 & -- & 53.50 & 77.57 & 79.12 & 75.51 & 80.74 & 72.93 & 73.23 \\
\cmidrule(lr){2-10}
& FlatQuant & RTN & 50.00 & 75.80 & 76.80 & 72.91 & 79.16 & 72.69 & 71.23 \\
& FlatQuant + \name & RTN
& 50.77 & 77.31 & 77.43 & 74.91 & 80.03 & 73.40 & \textbf{72.31}$^{*}$ \\
& FlatQuant & GPTQ & 50.51 & 75.88 & 76.49 & 73.20 & 79.00 & 72.93 & 71.33 \\
& FlatQuant + \name & GPTQ
& 49.83 & 77.15 & 77.47 & 74.64 & 79.87 & 72.45 & 71.90 \\

\midrule

\multirow{5}{*}{3-70B}
& FP16 & -- & 64.25 & 85.94 & 84.93 & 79.37 & 84.44 & 80.74 & 79.95 \\
\cmidrule(lr){2-10}
& FlatQuant & RTN & 62.12 & 84.97 & 83.95 & 78.73 & 84.28 & 80.03 & 79.01 \\
& FlatQuant + \name & RTN
& 63.40 & 84.09 & 84.37 & 79.20 & 83.30 & 80.03 & \textbf{79.07}$^{*}$ \\
& FlatQuant & GPTQ & 61.95 & 84.47 & 83.87 & 77.99 & 83.95 & 79.24 & 78.58 \\
& FlatQuant + \name & GPTQ
& 61.52 & 85.06 & 83.61 & 78.56 & 83.57 & 79.56 & 78.64 \\

\bottomrule
\end{tabular}%
}

\vspace{0.3em}
\small\textit{Note.} Higher accuracy is better.
``*'' indicates statistically significant improvements (\ie, two-sided $t$-test with $p<0.05$) over the best baseline.
\end{table*}

\subsection{Zero-shot QA Performance}
Table~\ref{tab:qa_4bit_phasequant_flatquant} reports the zero-shot QA results of LLaMA models using FlatQuant as the PTQ baseline under W4A4KV4 quantization. We focus on FlatQuant because it represents the state-of-the-art PTQ baseline under this setting. Therefore, improvements over FlatQuant provide stronger evidence that \textbf{\name} is complementary to existing PTQ transformations, rather than merely compensating for a weak baseline. Unlike perplexity, which mainly measures token-level language modeling quality, zero-shot QA more directly reflects downstream semantic understanding, commonsense reasoning, and context-based answer selection. This evaluation therefore examines whether the benefits of \textbf{\name} can transfer from language modeling metrics to task-level performance. Overall, \textbf{\name} improves the average QA score in most settings while keeping weights and KV cache at 4-bit precision and using 8-bit activations only at activation-sensitive layers.

The gains are more evident on smaller or more fragile quantized models. For LLaMA-2-7B, \textbf{\name} improves the average score from $67.96$ to $68.46$ under RTN and from $67.47$ to $68.45$ under GPTQ. A similar trend appears on LLaMA-3-8B, where the RTN setting improves from $71.23$ to $72.31$, giving one of the largest average gains in the table. These results suggest that when the quantized baseline is less stable, task-relevant representations are more vulnerable to local activation quantization errors and thus benefit more from phase-aware activation precision. In contrast, larger models such as LLaMA-2-70B and LLaMA-3-70B already retain strong zero-shot performance after FlatQuant, so their absolute gains are smaller. For example, LLaMA-2-70B improves from $76.62$ to $76.82$ under RTN and from $76.53$ to $76.62$ under GPTQ, while the average improvement on LLaMA-3-70B is more limited. This pattern shows that \textbf{\name} does not bring the same magnitude of improvement across all model scales; instead, it is most effective when the activation-side quantization bottleneck is more severe and the model representation is closer to its stability limit.

Another observation is that the improvements are not uniform across QA tasks. Tasks such as ARC-Easy, HellaSwag, and LAMBADA often benefit from \textbf{\name}, especially on smaller models, suggesting that protecting activation-sensitive layers helps preserve contextual understanding, sentence-level completion, and commonsense selection ability. By contrast, tasks such as WinoGrande or PIQA sometimes show smaller gains or slight drops. This differs from the perplexity results: perplexity is a continuous token-level metric and usually reflects quantization error more smoothly, whereas zero-shot QA accuracy is a discrete task-level metric that is more affected by answer choices, prompting format, and decision boundaries. Therefore, improving local activation precision may not always translate into consistent accuracy gains on every benchmark, but the average score better captures the overall trend. Taken together, these results show that \textbf{\name} not only reduces language modeling perplexity, but also improves downstream task robustness without changing the 4-bit weight and KV-cache configuration, with the clearest benefits appearing on smaller models and more fragile quantization settings.

\begin{figure*}[t]
    \centering
    \includegraphics[width=0.72\textwidth]{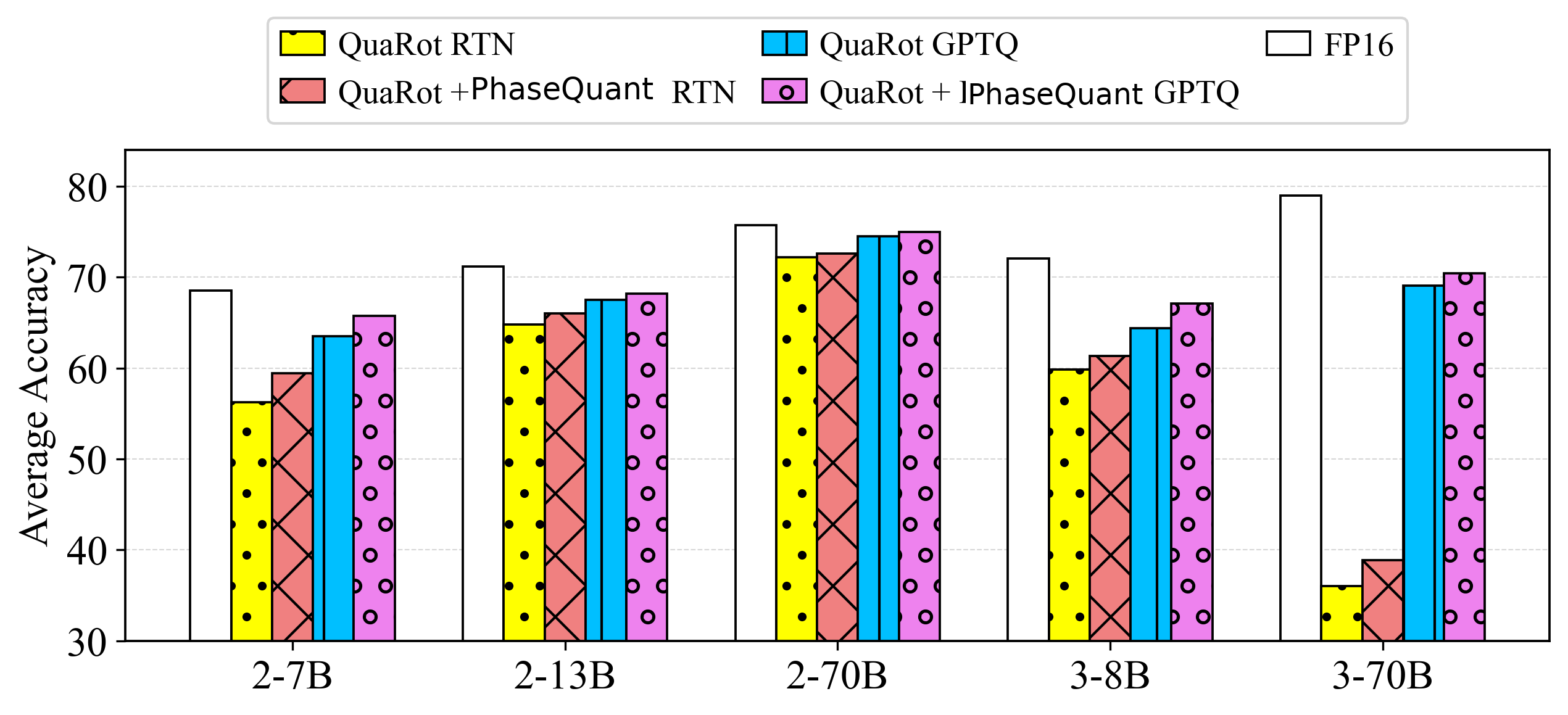}
    \vspace{-0.5em}
    \caption{Average zero-shot QA accuracy of QuaRot with and without \textbf{\name} under W4A4KV4 quantization.}
    \label{fig:quarot_avg_accuracy}
    \vspace{-0.8em}
\end{figure*}

\subsection{Generalization to Additional PTQ Backbones}
We evaluate \textbf{\name} on QuaRot under W4A4KV4 quantization, measuring average zero-shot QA accuracy on ARC-C, ARC-E, LAMBADA, PIQA, and WinoGrande. \textbf{\name} is evaluated with QuaRot without altering the weight quantizer or KV-cache precision. It follows a phase-aware mixed-precision activation policy, assigning 8-bit activations only to quantization-sensitive layers while keeping the remaining layers at 4-bit. The specific layer selections for each model follow the implementation described in Section~\ref{subsec:implementation_details}.

As illustrated in Figure~\ref{fig:quarot_avg_accuracy}, \textbf{\name} consistently enhances QuaRot across both RTN and GPTQ weight quantization settings. The improvement is most pronounced for smaller models (LLaMA-2-7B, LLaMA-2-13B, LLaMA-3-8B) but remains clearly positive for larger models (LLaMA-2-70B, LLaMA-3-70B). Notably, even in cases where RTN underperforms on LLaMA-3-70B, \textbf{\name} helps mitigate the degradation caused by low-bit quantization. Under GPTQ, where the baseline is already stronger, \textbf{\name} still delivers consistent gains. These results suggest that \textbf{\name} operates independently of the PTQ backbone or weight quantizer and effectively strengthens low-bit inference robustness by selectively adjusting activation precision across layers.

\subsection{Residual Dynamics in MoE Architectures}
\label{sec:moe_residual_dynamics}
FlatQuant reports that, compared with standard dense LLMs, Mixture-of-Experts (MoE) models such as DeepSeek-V3-Base and DeepSeek-R1 can still maintain performance close to their original 16-bit versions under W4A4 quantization. We explain this observation from the residual-stream perspective by analyzing DeepSeek-V2-Lite, a representative MoE model, and comparing it with the dense LLaMA-3-8B model.

For an MoE layer with top-$k$ routing, the token-wise update can be written as
\begin{equation}
\Delta \boldsymbol{x}^{(\ell)}_t
=
\sum_{e \in \mathcal{E}_t}
g^{(\ell)}_{t,e}
f^{(\ell)}_e(\boldsymbol{x}^{(\ell)}_t),
\end{equation}
where $\mathcal{E}_t$ denotes the selected experts for token $t$,
$g^{(\ell)}_{t,e}\geq 0$ is the routing weight, and
$f^{(\ell)}_e(\cdot)$ denotes the transformation of expert $e$.
By the triangle inequality, we have
\begin{equation}
\left\|\Delta \boldsymbol{x}^{(\ell)}_t\right\|_2
\leq
\sum_{e \in \mathcal{E}_t}
g^{(\ell)}_{t,e}
\left\|
f^{(\ell)}_e(\boldsymbol{x}^{(\ell)}_t)
\right\|_2 .
\end{equation}

This analysis suggests that, in an MoE layer, residual updates are routed through a small subset of experts and weighted by the router scores. When no single expert dominates the routing, the current update is less likely to overwhelm the historical residual state. Consequently, MoE layers exhibit smaller relative residual updates and better preservation of historical information under 4-bit activation quantization.

\begin{figure}[t]
    \centering
    \includegraphics[width=0.70\linewidth]{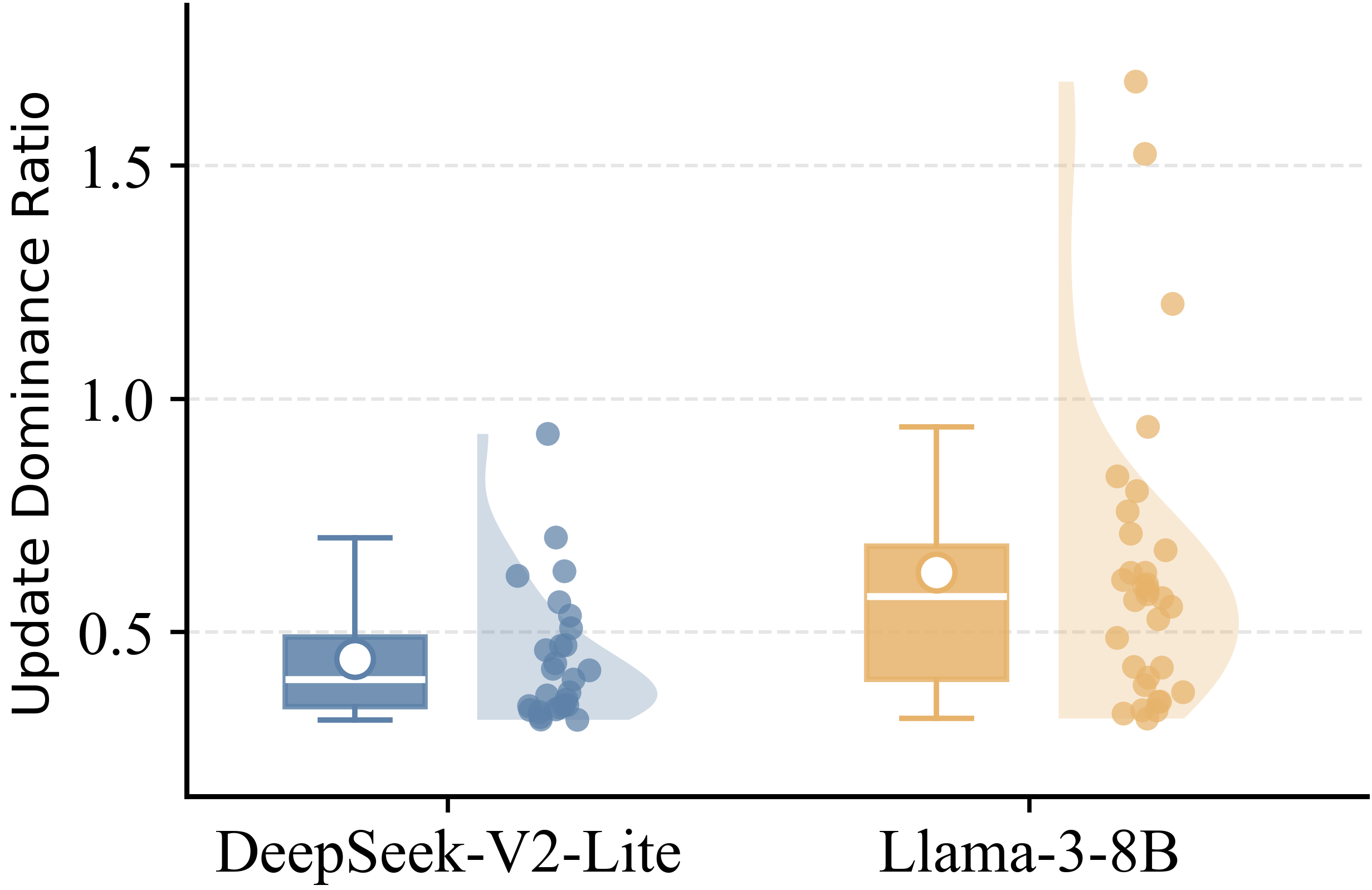}

    \vspace{1.2em}

    \includegraphics[width=0.70\linewidth]{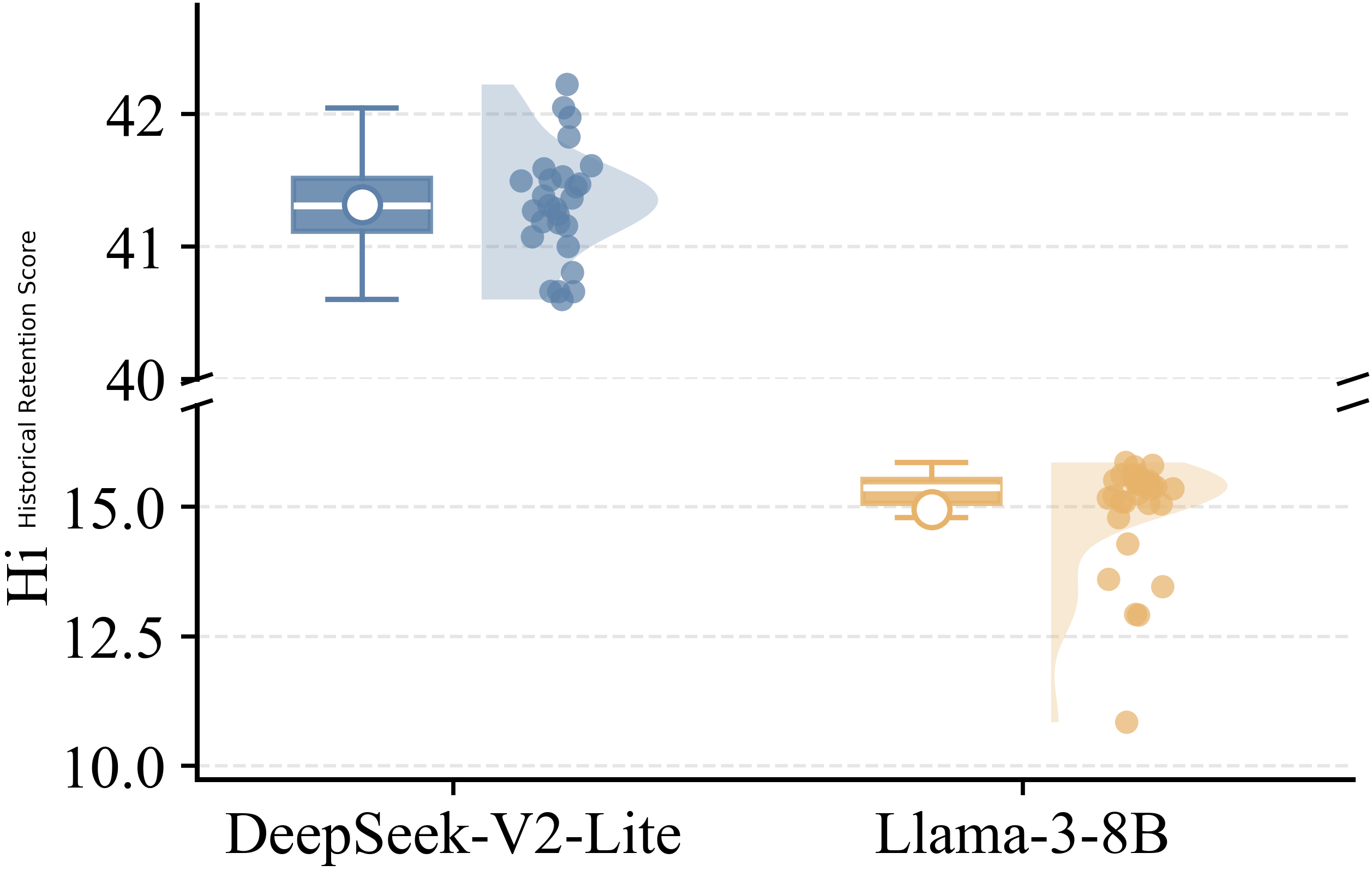}

    \vspace{-0.2em}
    \caption{Residual-stream dynamics of DeepSeek-V2-Lite and LLaMA-3-8B under 4-bit activation quantization, measured by the Update Dominance Ratio (UDR) and the Historical Retention Score (HRS).}
    \label{fig:moe_dense_dynamics}
    \vspace{-0.5em}
\end{figure}

Figure~\ref{fig:moe_dense_dynamics} confirms this behavior. Compared with the dense LLaMA-3-8B model, DeepSeek-V2-Lite shows a tighter distribution of UDR values and substantially higher HRS, indicating more stable updates and stronger retention of historical semantic features. The 4-bit quantizer thus rarely needs to expand its effective range to accommodate large updates, allowing smaller historical features to maintain sufficient quantization resolution. These findings offer a dynamic explanation for why MoE models handle low-bit PTQ more effectively: sparse expert routing smooths residual updates and helps retain historical information. In sum, our results highlight that the robustness of activation quantization is closely tied to residual-stream dynamics, rather than depending solely on static, layer-wise activation statistics.

\begin{figure*}[t]
    \centering
    \begin{subfigure}[t]{0.32\textwidth}
        \centering
        \includegraphics[width=\linewidth]{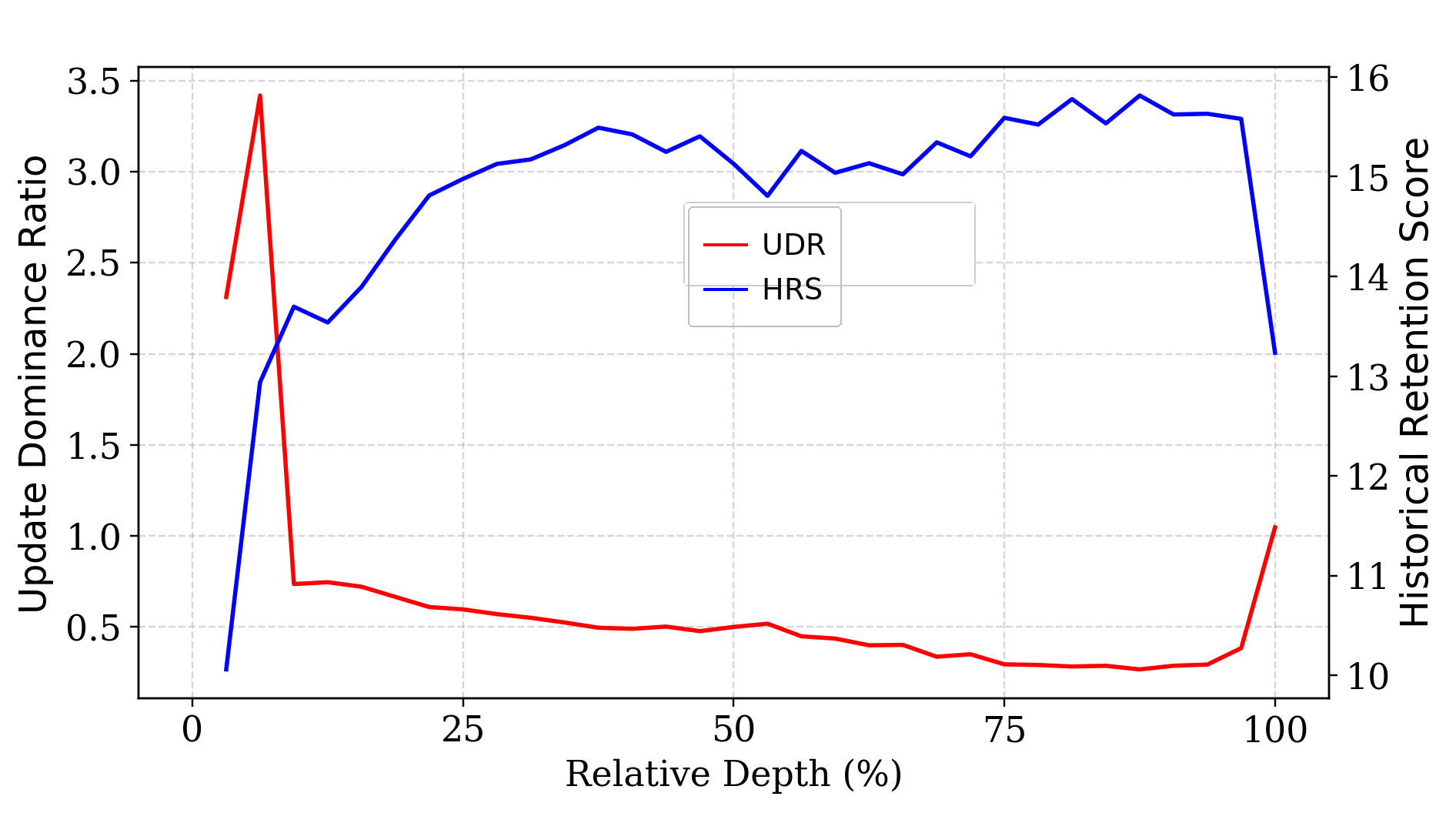}
        \caption{LLaMA-2-7B}
        \label{fig:layerwise_7b}
    \end{subfigure}
    \hfill
    \begin{subfigure}[t]{0.32\textwidth}
        \centering
        \includegraphics[width=\linewidth]{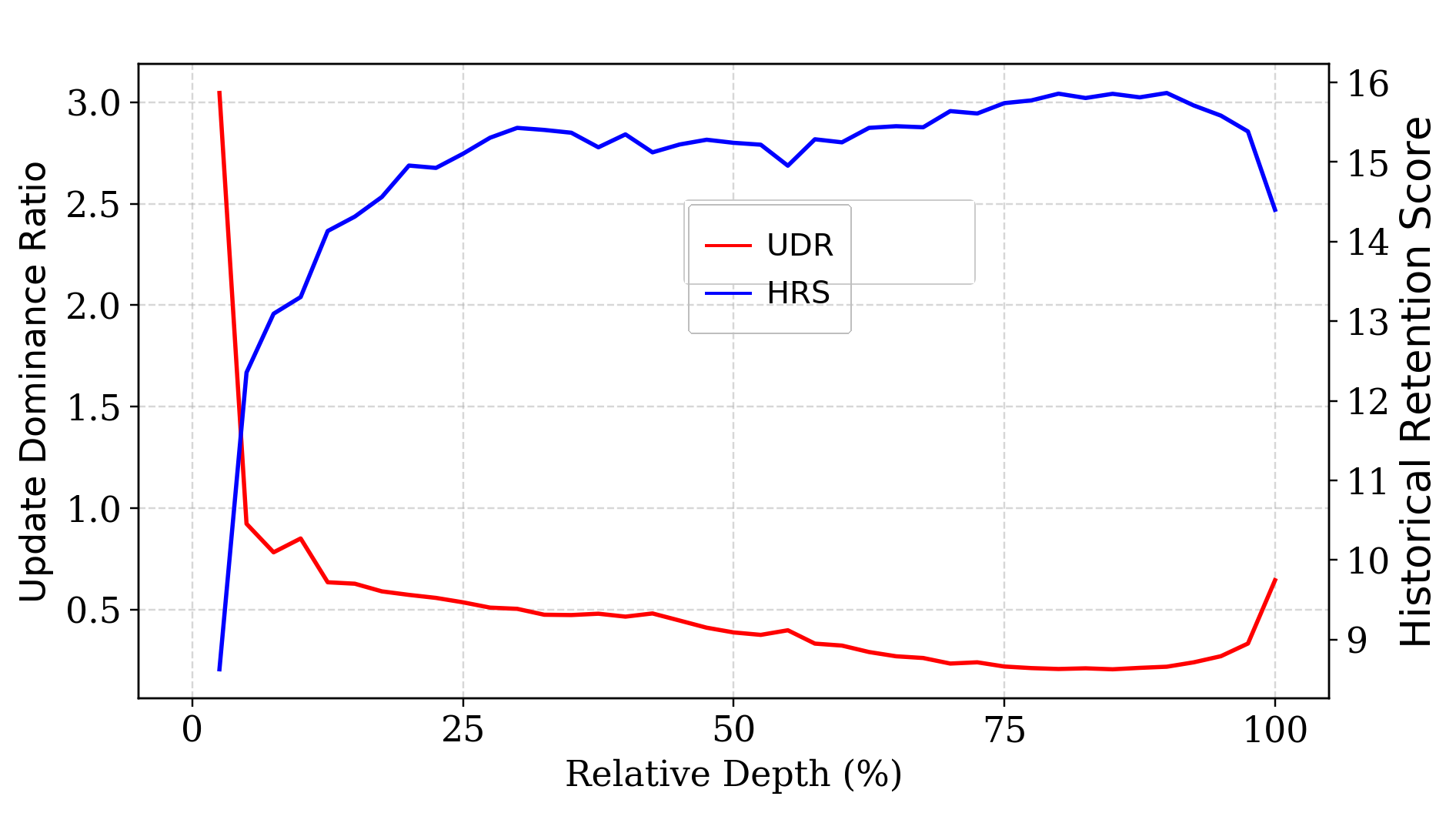}
        \caption{LLaMA-2-13B}
        \label{fig:layerwise_13b}
    \end{subfigure}
    \hfill
    \begin{subfigure}[t]{0.32\textwidth}
        \centering
        \includegraphics[width=\linewidth]{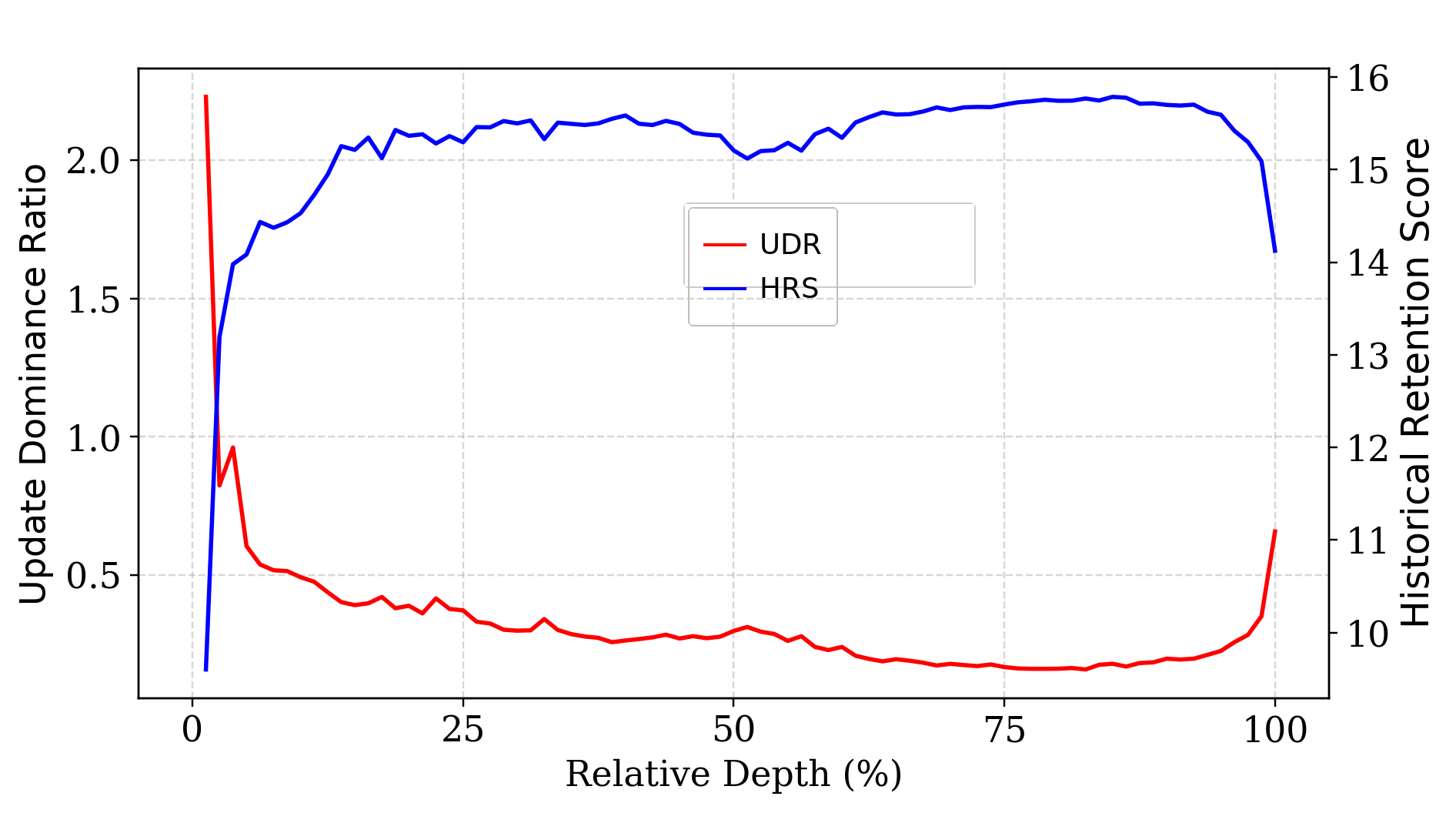}
        \caption{LLaMA-2-70B}
        \label{fig:layerwise_70b}
    \end{subfigure}
    \vspace{-0.5em}
    \caption{Layer-wise residual-stream dynamics across LLaMA-2 models, measured by the UDR (left axis) and the HRS (right axis).}
    \label{fig:layerwise_llama2}
    \vspace{-0.8em}
\end{figure*}

\begin{figure*}[t]
    \centering
    \begin{subfigure}[t]{0.32\textwidth}
        \centering
        \includegraphics[width=\linewidth]{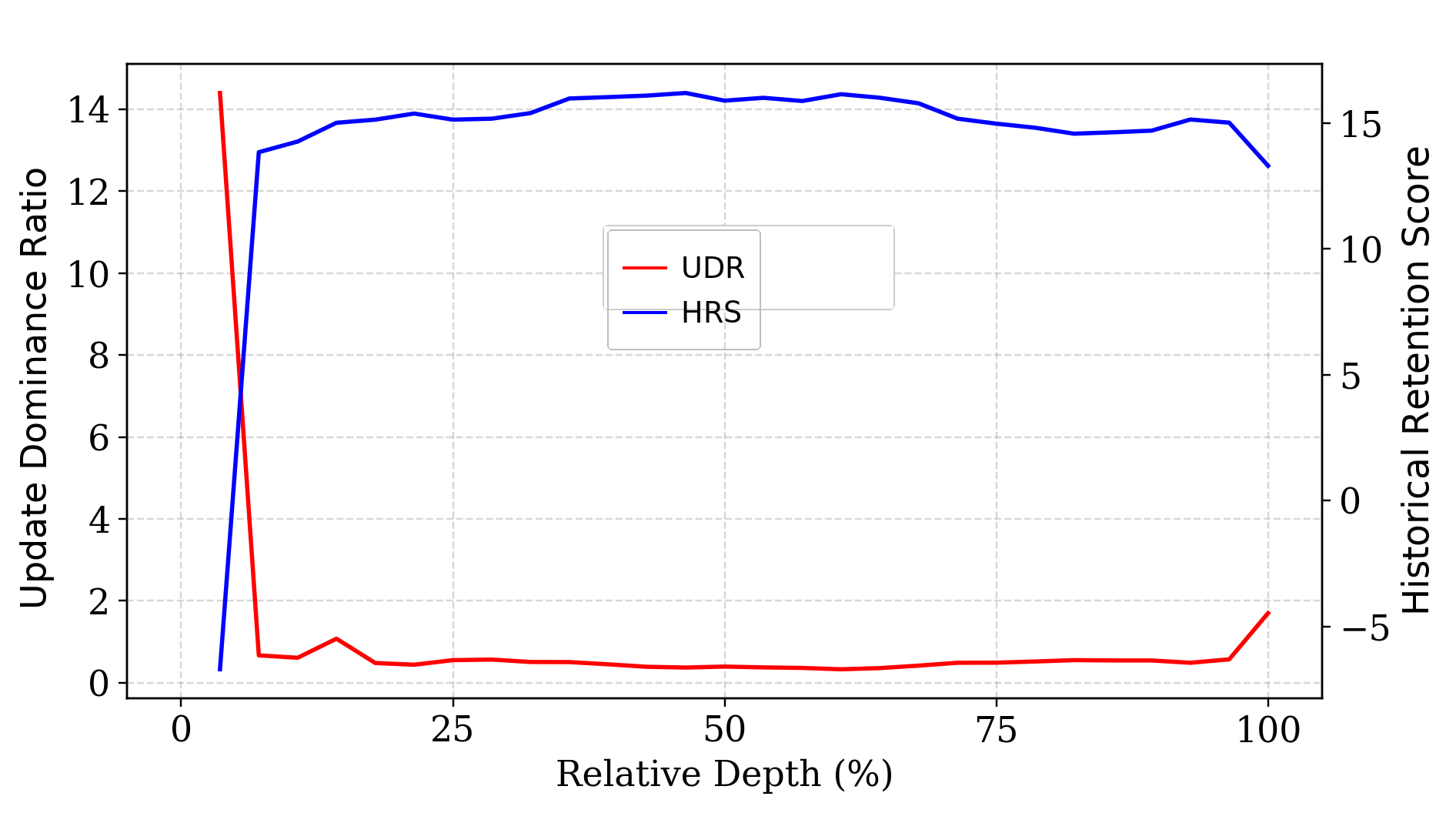}
        \caption{Qwen2.5-7B}
        \label{fig:layerwise_qwen}
    \end{subfigure}
    \hfill
    \begin{subfigure}[t]{0.32\textwidth}
        \centering
        \includegraphics[width=\linewidth]{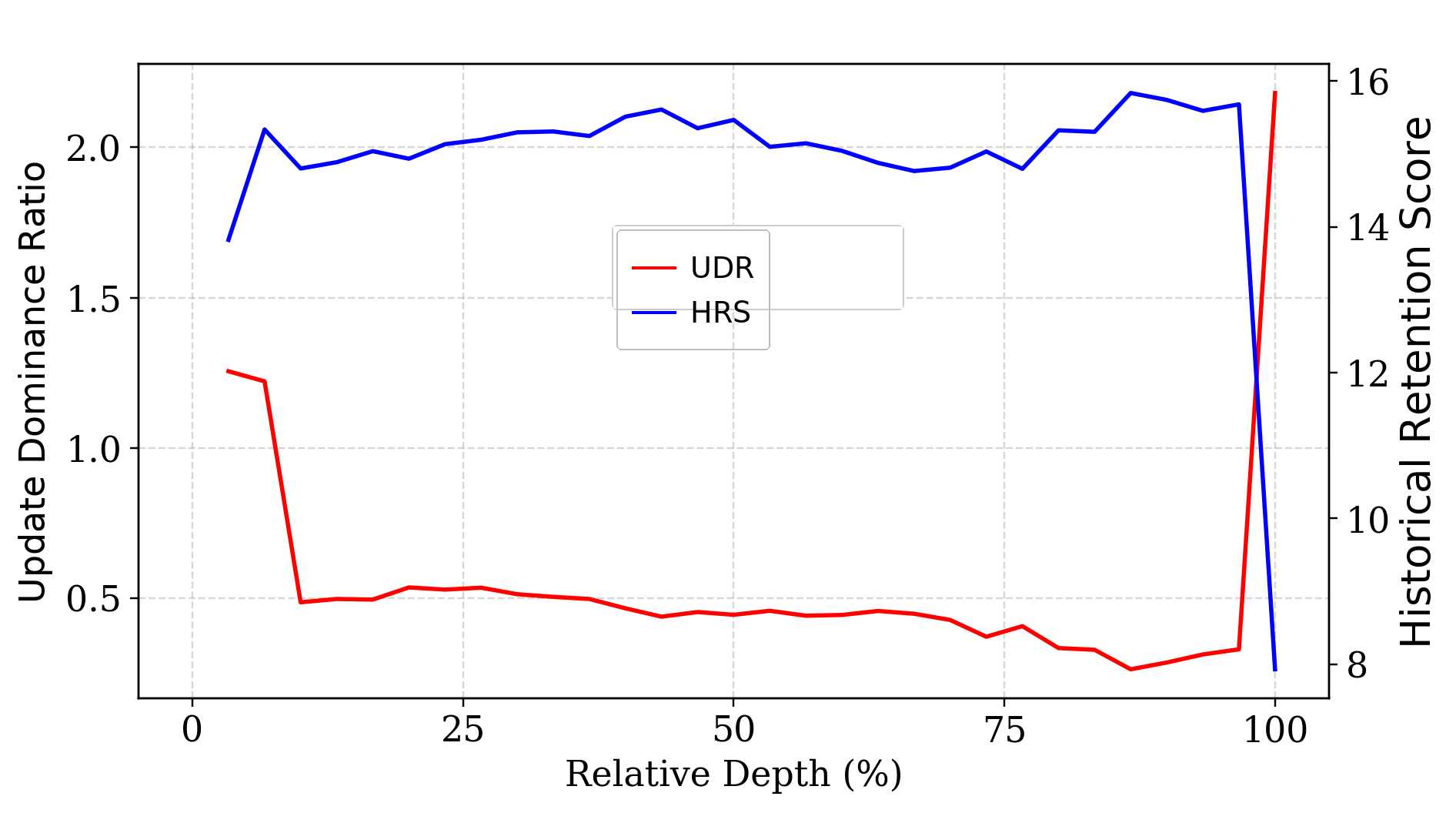}
        \caption{DeepSeek-7B}
        \label{fig:layerwise_deepseek}
    \end{subfigure}
    \hfill
    \begin{subfigure}[t]{0.32\textwidth}
        \centering
        \includegraphics[width=\linewidth]{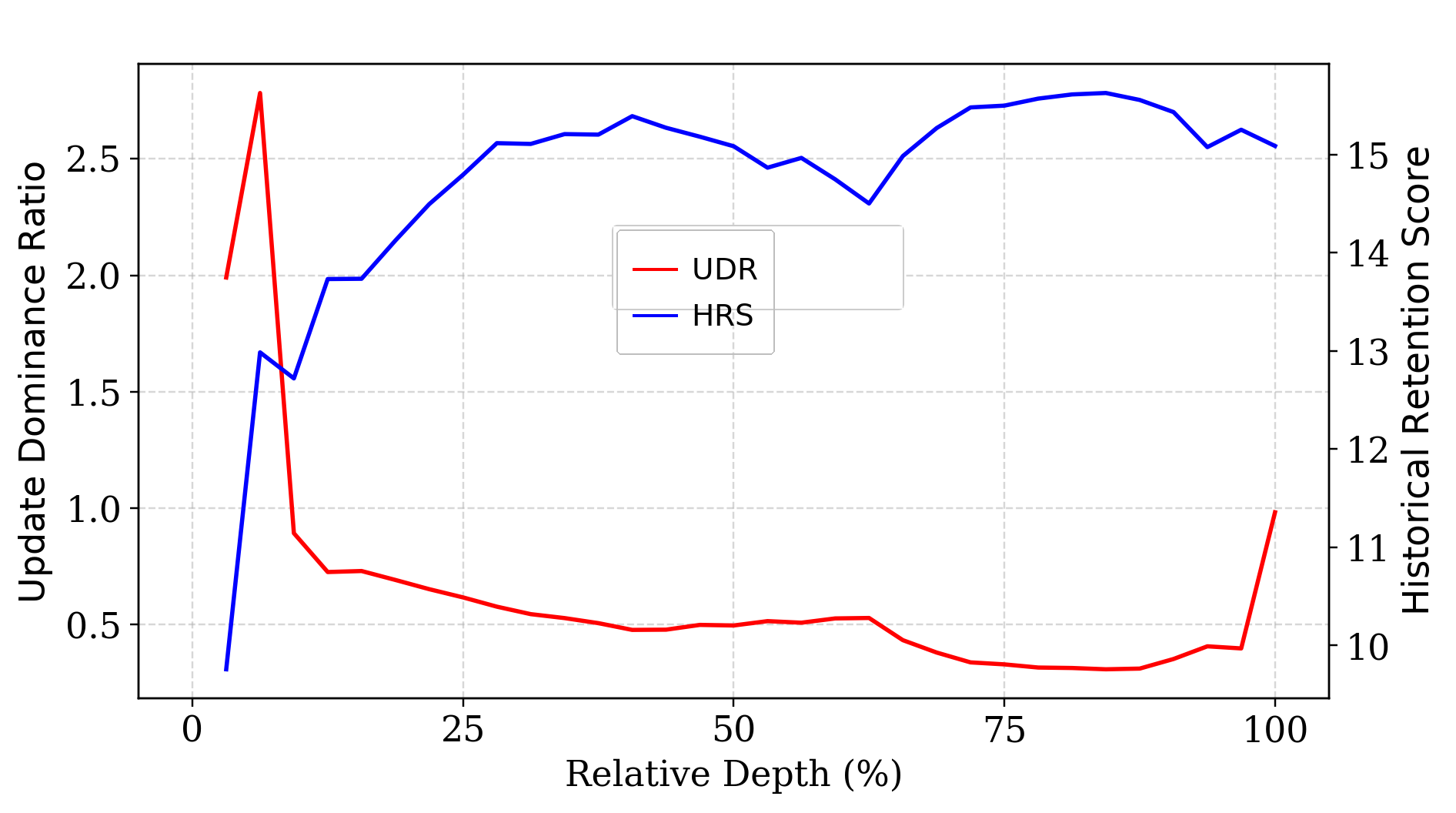}
        \caption{Mistral-7B}
        \label{fig:layerwise_mistral}
    \end{subfigure}
    \vspace{-0.5em}
    \caption{Layer-wise residual-stream dynamics across different 7B-scale decoder-only model families, measured by the UDR (left axis) and the HRS (right axis).}
    \label{fig:layerwise_three_models}
    \vspace{-0.8em}
\end{figure*}

\subsection{Phase-wise Residual-Stream Dynamics Across Dense LLMs}
We evaluate the phase-wise residual-stream dynamics across dense LLMs of different parameter scales and model families under 4-bit activation-only quantization, using sequences constructed from WikiText-2 (20 samples, sequence length 512) and applying a FlatQuant activation-only quantization pipeline. Figure~\ref{fig:layerwise_llama2} compares LLaMA-2 models with varying sizes, while Figure~\ref{fig:layerwise_three_models} compares 7B-scale models from Qwen, DeepSeek, and Mistral series~\cite{yang2025qwen3,guo2025deepseek,jiang2023mistral7b}. Across all models, a consistent pattern emerges: the appearance and disappearance of massive activations coincide with higher UDR and lower HRS, whereas middle layers remain relatively stable. In particular, for the DeepSeek model, the last layers exhibit a steep increase in residual updates and a sharp drop in HRS, indicating that these final layers are especially sensitive to 4-bit activation-only quantization. While larger models generally show smoother residual dynamics overall, the phase boundaries remain clearly visible across scales and families. These observations further support the use of phase-aware activation precision in \textbf{\name}.

\subsection{Additional Results on Complex Reasoning Tasks}
\label{sec:reasoning_1_5b}

Recent work has investigated how post-training quantization affects complex
reasoning models~\cite{liu2025quantization}. Following this line of evaluation,
we further examine whether \textbf{\name} can improve reasoning performance under
aggressive low-bit quantization. Specifically, we evaluate
\textbf{\name} on DeepSeek-R1-Distill-Qwen-1.5B~\cite{guo2025deepseek}, a compact reasoning model
distilled from DeepSeek-R1.

We evaluate on four reasoning benchmarks: MATH-500~\cite{hendrycks2021measuring},
GSM8K~\cite{cobbe2021training}, GPQA-Diamond~\cite{rein2023gpqa}, and
LiveCodeBench~\cite{jain2025livecodebench}. These benchmarks cover advanced
mathematical reasoning, grade-school mathematical reasoning, graduate-level scientific reasoning, and code-related reasoning, respectively. Compared with the implementation details in
Section~\ref{subsec:implementation_details}, we increase the FlatQuant calibration
sequence length to 4096 to better match the long-context reasoning setting. For
\textbf{\name}, we apply 8-bit activation precision to layers
$\{1,2,3,28\}$ and keep the remaining activations, weights, and KV cache at
4-bit precision. As shown in Table~\ref{tab:deepseek_qwen_1_5b_reasoning},
\textbf{\name} consistently improves FlatQuant across all four benchmarks,
raising the average score from $43.58$ to $47.36$. The largest gain appears on
GPQA-Diamond, indicating that phase-aware activation precision is particularly
beneficial for challenging reasoning tasks sensitive to low-bit activation
quantization error.

\begin{table}[t]
\centering
\small
\caption{Reasoning performance of DeepSeek-R1-Distill-Qwen-1.5B under W4A4KV4 quantization.}
\label{tab:deepseek_qwen_1_5b_reasoning}
\vspace{-0.5em}
\setlength{\tabcolsep}{4pt}
\renewcommand{\arraystretch}{1.08}
\begin{tabular}{lccc}
\toprule
\textbf{Benchmark}
& \textbf{16-bit}
& \textbf{FlatQuant}
& \textbf{+\name} \\
\midrule
MATH-500       & 84.40 & 63.20 & 67.40 \\
GSM8K          & 84.61 & 78.62 & 79.61 \\
GPQA-Diamond   & 36.87 & 25.76 & 33.84 \\
LiveCodeBench  & 16.04 & 6.72  & 8.58  \\
\midrule
Average        & 55.48 & 43.58 & 47.36 \\
\bottomrule
\end{tabular}
\vspace{-0.7em}
\end{table}

\subsection{Efficiency Analysis}
We evaluate Meta-Llama-3-8B under the W4A4KV4 setting on a single NVIDIA RTX PRO 6000 Blackwell Server Edition GPU, with a prefill length of 2048 tokens and 256 decoding steps across different batch sizes. FlatQuant keeps all activations at 4-bit precision, whereas \textbf{\name} uses 8-bit activations only for the four quantization-sensitive layers of the 32-layer model, \ie, layers $1$--$3$ and the final layer $32$.
All results are averaged over five runs after two warmup runs.

As shown in Tables~\ref{tab:prefill_throughput_memory} and~\ref{tab:decode_throughput_memory}, \textbf{\name} improves throughput in both prefill and decode stages without introducing runtime slowdown.
In the prefill stage, throughput consistently increases by $1.05\times$--$1.06\times$, while in the decode stage the improvement remains at $1.05\times$--$1.07\times$ across different batch sizes.
The additional memory cost is also limited.
Although assigning 8-bit activations to selected quantization-sensitive layers increases peak memory at small batch sizes, the relative overhead becomes smaller as the batch size grows.
For prefill, the memory ratio decreases from $1.18\times$ at batch size 1 to $1.02\times$ at batch size 32; for decode, it decreases from $1.20\times$ to $1.04\times$.
These results show that \textbf{\name} preserves practical inference efficiency, with a modest memory overhead that becomes smaller at larger batch sizes.

\begin{table}[t]
\centering
\small
\caption{Prefill efficiency of Llama-3-8B under W4A4KV4 quantization.}
\label{tab:prefill_throughput_memory}
\vspace{-0.5em}
\setlength{\tabcolsep}{2.8pt}
\renewcommand{\arraystretch}{1.05}
\resizebox{\columnwidth}{!}{%
\begin{tabular}{c c c c c c c}
\toprule
\multirow{2}{*}{\textbf{Batch}}
& \multicolumn{3}{c}{\textbf{Throughput (tok/s)}}
& \multicolumn{3}{c}{\textbf{Peak Memory (GB)}} \\
\cmidrule(lr){2-4}
\cmidrule(lr){5-7}
& {\scriptsize \textbf{FlatQuant}}
& {\scriptsize \textbf{+\name}}
& {\scriptsize \textbf{Ratio}}
& {\scriptsize \textbf{FlatQuant}}
& {\scriptsize \textbf{+\name}}
& {\scriptsize \textbf{Ratio}} \\
\midrule
1  & 3444.25 & 3660.80 & 1.06$\times$ & 6.97  & 8.19  & 1.18$\times$ \\
4  & 4072.05 & 4312.93 & 1.06$\times$ & 12.17 & 13.39 & 1.10$\times$ \\
16 & 4231.78 & 4448.35 & 1.05$\times$ & 32.97 & 34.19 & 1.04$\times$ \\
32 & 4246.06 & 4462.65 & 1.05$\times$ & 60.70 & 61.92 & 1.02$\times$ \\
\bottomrule
\end{tabular}}

\vspace{-0.8em}
\end{table}

\begin{table}[t]
\centering
\small
\caption{Decode efficiency of Llama-3-8B under W4A4KV4 quantization.}
\label{tab:decode_throughput_memory}
\vspace{-0.5em}
\setlength{\tabcolsep}{2.8pt}
\renewcommand{\arraystretch}{1.05}
\resizebox{\columnwidth}{!}{%
\begin{tabular}{c c c c c c c}
\toprule
\multirow{2}{*}{\textbf{Batch}}
& \multicolumn{3}{c}{\textbf{Throughput (tok/s)}}
& \multicolumn{3}{c}{\textbf{Peak Memory (GB)}} \\
\cmidrule(lr){2-4}
\cmidrule(lr){5-7}
& {\scriptsize \textbf{FlatQuant}}
& {\scriptsize \textbf{+\name}}
& {\scriptsize \textbf{Ratio}}
& {\scriptsize \textbf{FlatQuant}}
& {\scriptsize \textbf{+\name}}
& {\scriptsize \textbf{Ratio}} \\
\midrule
1  & 4.66   & 4.98   & 1.07$\times$ & 6.08  & 7.30  & 1.20$\times$ \\
4  & 18.61  & 19.81  & 1.06$\times$ & 8.63  & 9.85  & 1.14$\times$ \\
16 & 65.37  & 69.08  & 1.06$\times$ & 18.81 & 20.03 & 1.06$\times$ \\
32 & 114.48 & 120.16 & 1.05$\times$ & 32.39 & 33.61 & 1.04$\times$ \\
\bottomrule
\end{tabular}}

\vspace{-0.8em}
\end{table}

\section{Related Work}
\label{sec:related_work}

\subsection{Post-training Quantization}
Post-training quantization (PTQ) is widely used to reduce the memory footprint and inference cost of large language models without full-model retraining. Early PTQ methods mainly focus on weight-only quantization, in which weights are compressed to low-bit formats while activations remain at higher precision. Representative methods such as GPTQ~\cite{frantar2022gptq} and AWQ~\cite{lin2024awq} reduce weight quantization error through approximate second-order reconstruction or activation-aware channel protection, while ZeroQuant~\cite{yao2022zeroquant} and OmniQuant~\cite{shao2024omniquant} improve low-bit quantization through fine-grained quantization, learnable clipping, and equivalent transformations. However, weight-only quantization provides limited inference acceleration and memory reduction when activations and KV caches remain in high precision. Recent work, therefore, moves toward weight-activation and KV-cache quantization under W4A4 or W4A4KV4 settings~\cite{ashkboos2024quarot,liu2025spinquant,sun2024flatquant}. Compared with weights, activations are more difficult to quantize because they are input- and token-dependent and highly non-stationary across layers.

In contrast to prior PTQ methods that mainly optimize local layer-wise reconstruction, our work studies why 4-bit activation errors concentrate in specific depth regions and proposes a residual-dynamics-aware precision allocation strategy.

\subsection{Massive Activations and Smoothing}
A major challenge in low-bit activation quantization is the presence of massive activations, where extremely large values concentrate on a small number of tokens and feature dimensions~\cite{xiao2023smoothquant,dettmers2022gpt3,sun2024massive}. These activations are often associated with special pivot tokens, such as the BOS token, which dominate the token-wise dynamic range, making 4-bit quantization insufficient to preserve ordinary hidden features. To mitigate this issue, existing PTQ methods commonly apply pre-quantization distribution smoothing. SmoothQuant~\cite{xiao2023smoothquant} uses per-channel scaling to migrate quantization difficulty from activations to weights. QuaRot~\cite{ashkboos2024quarot} applies randomized Hadamard rotations to spread outlier energy across hidden dimensions, while SpinQuant~\cite{liu2025spinquant} learns optimized orthogonal rotations to reduce the instability of random rotations. FlatQuant~\cite{sun2024flatquant} further introduces learnable affine transformations to improve the flatness of weight and activation distributions. Despite their effectiveness, these methods largely follow a static layer-wise view: each Transformer block is calibrated with a fixed transformation and quantized in the same way during inference. They mainly address spatial distribution imbalance, such as channel-wise outliers or local activation sharpness, but do not explicitly model how the residual stream evolves across depth. In contrast, \name{} focuses on cross-layer quantization instability. Since orthogonal transformations preserve Euclidean norms, they cannot reduce the norm imbalance between newly injected updates and historical residual information. Near phase-boundary layers, 4-bit quantization may therefore still fail to preserve smaller but important historical semantic features even after distribution smoothing.

\subsection{Activation Precision under Residual Stream Dynamics}
Recent studies suggest that massive activations are not merely static
channel-wise outliers, but are also closely related to residual-stream dynamics
across network depth. For example, attention sinks and compression valleys have
been associated with massive activations in the residual stream, where special
tokens such as the BOS token can accumulate large residual norms, introduce
input-agnostic representation biases, and form phase-transition-like patterns in
internal information flow~\cite{queipo2025attention}. This perspective indicates
that low-bit activation degradation may also arise from the interaction between
newly injected layer-wise updates and accumulated historical residual
information. Mixed-precision quantization has been widely explored to avoid
assigning the same precision to all model components. LLM.int8()~\cite{dettmers2022gpt3}
isolates emergent outlier feature dimensions and computes them in 16-bit
precision while keeping most values in 8-bit matrix multiplication.
Atom~\cite{zhao2024atom} combines mixed precision, fine-grained group
quantization, dynamic activation quantization, KV-cache quantization, and kernel
co-design for efficient W4A4 serving. For long-context inference,
KIVI~\cite{liu2024kivi} applies per-channel quantization to key caches and
per-token quantization to value caches, while KVQuant~\cite{hooper2024kvquant}
further improves KV-cache quantization with sensitivity-aware datatypes and
dense-and-sparse outlier handling. However, these methods mainly protect outlier
dimensions, sensitive components, or KV-cache entries based on distributional
statistics, hardware efficiency, or empirical sensitivity, without explicitly
examining whether historical residual information is preserved after low-bit
activation quantization. \name{} addresses this gap from a residual-dynamics
perspective by introducing the Update Dominance Ratio and the Historical
Retention Score to identify layers where 4-bit activation quantization becomes
unstable due to abrupt residual updates and degraded historical feature
preservation. It then applies 8-bit activation precision only to these sensitive
layers while keeping the remaining layers in 4-bit. In this sense, \name{} is
complementary to existing transformation-based PTQ methods: prior methods mainly
smooth activation distributions within the layer-wise feature space, whereas
\name{} targets dynamic instability along the network-depth dimension.

\section{Conclusion}
\label{sec:conclusion}

In this paper, we propose \textbf{\name}, a residual-stream-dynamics-aware
mixed-precision activation policy for efficient low-bit LLM inference. Using the
Update Dominance Ratio and the Historical Retention Score, \textbf{\name}
identifies quantization-sensitive layers and assigns 8-bit activation precision
only to them, while keeping weights, KV caches, and the remaining activations at
4-bit precision.
Experiments show that \textbf{\name} consistently improves strong PTQ baselines
across model families, datasets, and weight quantizers. It reduces perplexity,
improves zero-shot QA performance in most settings, and maintains practical
inference efficiency with modest memory overhead. These results suggest that
low-bit activation robustness depends not only on static layer-wise activation
distributions, but also on cross-layer residual-stream dynamics.

\paragraph{\textbf{Limitations and Future Work.}}
In practical deployment, PTQ methods may still suffer from unstable performance
due to calibration data mismatch or model-specific activation behaviors, making
them less reliable for accuracy-sensitive applications. This issue becomes more
pronounced under aggressive low-bit settings, where a small number of
quantization-sensitive layers can disproportionately affect the final model
quality. \textbf{\name} mitigates this limitation by analyzing residual-stream
dynamics and selectively assigning higher activation precision to these
sensitive layers, while avoiding a uniform increase in activation precision.
Future work may extend this direction toward finer-grained layer selection and
token-level precision allocation for more adaptive low-bit inference.

\appendix
\section{Implementation Details and Hyperparameters}
\label{app:implementation}

Table~\ref{tab:hyperparameters} summarizes the calibration and quantization
hyperparameters shared by all methods, and Table~\ref{tab:sensitive_layers}
lists the phase-aware sensitive layer set $\mathcal{S}=\{1,2,3,L\}$ instantiated
for each evaluated model. All methods use the same calibration split, weight
quantizer, and KV-cache quantization protocol, so that the only difference
introduced by \textbf{\name} is the activation bit-width assignment across
layers.

\begin{table}[h]
\centering
\small
\caption{Shared calibration and quantization hyperparameters.}
\label{tab:hyperparameters}
\vspace{-0.5em}
\setlength{\tabcolsep}{4pt}
\renewcommand{\arraystretch}{1.1}
\resizebox{\columnwidth}{!}{%
\begin{tabular}{ll}
\toprule
\textbf{Item} & \textbf{Setting} \\
\midrule
Calibration source & WikiText-2 \\
Calibration length & 2048 tokens \\
Weight quantization & 4-bit, per-channel symmetric \\
Activation quantization & 4/8-bit, per-token symmetric \\
KV-cache quantization & 4-bit, group size 128 \\
FlatQuant training & 15 epochs, 128 sequences \\
FlatQuant batch size & 4 \\
LR (transformation params) & $5\times10^{-3}$ \\
LR (clipping params) & $5\times10^{-2}$ \\
Efficiency runs & 5 runs after 2 warmups \\
\bottomrule
\end{tabular}}
\end{table}

\begin{table}[h]
\centering
\small
\caption{Phase-aware sensitive layer sets $\mathcal{S}$ (layers indexed from $1$ to $L$).}
\label{tab:sensitive_layers}
\vspace{-0.5em}
\setlength{\tabcolsep}{6pt}
\renewcommand{\arraystretch}{1.1}
\resizebox{\columnwidth}{!}{%
\begin{tabular}{lcc}
\toprule
\textbf{Model} & \textbf{Depth $L$} & \textbf{8-bit activation layers $\mathcal{S}$} \\
\midrule
LLaMA-2-7B                 & 32 & $\{1,2,3,32\}$ \\
LLaMA-3-8B                 & 32 & $\{1,2,3,32\}$ \\
LLaMA-2-13B                & 40 & $\{1,2,3,40\}$ \\
LLaMA-2-70B                & 80 & $\{1,2,3,80\}$ \\
LLaMA-3-70B                & 80 & $\{1,2,3,80\}$ \\
DeepSeek-R1-Distill-Qwen-1.5B & 28 & $\{1,2,3,28\}$ \\
\bottomrule
\end{tabular}}
\end{table}

For the reasoning evaluation on DeepSeek-R1-Distill-Qwen-1.5B
(Section~\ref{sec:reasoning_1_5b}), the FlatQuant calibration sequence length is
increased to 4096 tokens to better match the long-context reasoning setting.

\section{Additional Reasoning Baselines}
\label{app:additional_reasoning}

Beyond the four reasoning benchmarks reported in
Table~\ref{tab:deepseek_qwen_1_5b_reasoning}, we also observe the same trend on
AIME-120, which contains competition-level mathematical problems: under W4A4KV4
quantization, the 16-bit DeepSeek-R1-Distill-Qwen-1.5B obtains $21.67$, while
FlatQuant drops to $10.00$. This severe degradation on the hardest mathematical
reasoning benchmark is consistent with our analysis that activation-side
quantization error disproportionately affects fragile, small distilled reasoning
models, and it further motivates phase-aware activation precision allocation for
such models.

\bibliographystyle{unsrt}
\bibliography{references}

\end{document}